%% file: main.tex
\documentclass[runningheads]{llncs}

\usepackage{accv}

\usepackage{accvabbrv}

\usepackage{graphicx}
\usepackage{booktabs}

\usepackage[accsupp]{axessibility}  
\usepackage{float}

\usepackage{hyperref}

\usepackage{orcidlink}

\begin{document}

\title{Direction-Aware Kinematics-Conditioned Graph Hypernetwork for Human Trajectory Prediction} 

\titlerunning{Kinematics-Conditioned Graph Hypernetwork for Trajectory Prediction}
\author{
Amirreza M. Shebly\inst{1}\orcidlink{0009-0001-4263-0217} \and
Mohammad Sarhangzadeh\inst{1}\orcidlink{0009-0004-3063-8141} \and
Doruk Oner\inst{1}\orcidlink{0000-0002-9403-4628}
}

\authorrunning{A.~M.~Shebly et al.}

\institute{
NeuraVision Lab, Department of Computer Engineering, Bilkent University, Ankara, Türkiye\\
\email{\{amirreza, m.sarhangzadeh, doruk.oner\}@bilkent.edu.tr}
}

\maketitle

\input{sec/0_abstract}
\input{sec/1_introduction}
\input{sec/2_related}
\input{sec/3_method}
\input{sec/4_experiments}
\input{sec/5_conclusion}

%
%
\bibliographystyle{splncs04}
\bibliography{main}
\end{document}



\title{Supplementary Material}

\titlerunning{Supplementary Material}

\author{}

\institute{}

\maketitle

\vspace{1em}

This supplementary material provides additional implementation
details and analyses of the proposed kinematics-conditioned interaction
modeling. We first describe the architecture, graph construction, loss terms,
training settings, and dataset-specific configurations used in the experiments.
We then provide diagnostic visualizations and quantitative analyses of the
learned interaction behavior, including directional anisotropy, temporal
consistency of the constructed interaction graphs, the dependence of FiLM
modulation weights on kinematic cues, additional qualitative and connectivity
analysis on NBA, and physical plausibility of the predicted trajectories.
These results provide further evidence for the design choices and empirical
behavior discussed in the main paper.

\section{Implementation Details}

\subsection{Architecture}
The temporal encoder embeds each agent's observed trajectory with a
multi-scale convolutional block (parallel kernels of size $3$ and $5$)
concatenated with per-agent kinematic statistics, projected to hidden
dimension $d\!=\!64$. The interaction backbone uses $L\!=\!3$ direction-aware
message-passing layers (Eqs.~(6)--(8)). Directional selectivity is realised by
three role heads whose Gaussian angular kernel (Eq.~(10)) has width
$\sigma_{\mathrm{ang}}=0.6$\,rad, combined through the FiLM modulation of Eqs.~(7),~(11),
and~(12). Dropout of $0.1$ is applied throughout. The full model has
around $0.16$M parameters.

\subsection{CVAE decoder}
Decoding uses a conditional VAE (Eq.~(13)). A prior network
$p_\theta(z\!\mid\!x)$ and a posterior network $q_\phi(z\!\mid\!x,y)$---each a
two-layer MLP over the interaction-refined node feature $\mathbf{h}_i$, the
posterior additionally conditioning on an encoding of the future---parameterise
diagonal-Gaussian latents of dimension $16$ (ETH/UCY default; per-dataset
values are listed in Table~\ref{tab:impl_perdataset}). Latents are drawn by
reparameterisation, $z=\mu+\sigma\odot\epsilon$, injected into the node feature
through $W_z$ and decoded to the predicted trajectory (Eq.~(13)). At inference
we draw $K\!=\!20$ latents from the prior and report best-of-$20$
minADE/minFDE.

\subsection{Kinematics-conditioned interaction graph}
Edges are gated by the four kinematic descriptors of Eqs.~(1)--(4): approach
rate (Eq.~(1)), inverse time-to-closest-approach score (Eq.~(2), clamped to $[-10,10]$), smoothed-direction alignment (Eq.~(3), with the motion direction an
$\alpha$-blend of the last two velocity estimates, $\alpha\!=\!0.7$), and distance attenuation (Eq.~(4),
scale $\sigma_d\!=\!2.0$\,m by default). In the interaction potential of
Eq.~(5), these four descriptors are combined with a forward-bearing term
$\max(\cos b_{ij},0)$, giving five weighted terms
$\boldsymbol{\lambda}\!=\!(\lambda_1,\dots,\lambda_5)\!=\!(1.0,0.5,1.0,1.0,0.5)$.
The potential is smoothed across frames by an exponential moving average
(factor $0.6$) and gated with hysteresis thresholds $\tau_{\text{on}}\!=\!0.5$
and $\tau_{\text{off}}\!=\!0.2$; each node retains at most $k\!=\!8$ neighbours
in $\mathcal{N}(i)$ by default. Per-dataset values of $\sigma_d$, $k$, and the
latent dimension are overrides of these ETH/UCY defaults, listed in
Table~\ref{tab:impl_perdataset}.

\subsection{Loss}
The probabilistic model is trained with the best-of-$K$ reconstruction loss of
Eq.~(15) (the deterministic variant uses Eq.~(14)), the KL term of Eq.~(16)
with weight $\beta_{\text{KL}}\!=\!1.0$ linearly annealed over the first $50$
epochs, and the three kinematic terms of Eqs.~(17),~(18), and~(19) with weights
$1.0$, $0.5$, and $0.1$, respectively.

\subsection{Training}
All models use Adam (learning rate $5\!\times\!10^{-3}$, weight decay
$5\!\times\!10^{-4}$), batch size $128$, and gradient clipping at $5.0$, with
random rotation augmentation and a fixed seed of $42$ for the primary experiments. Experiments are
implemented in PyTorch and run on a single NVIDIA RTX 6000 PRO GPU.

\subsection{Datasets and protocol}
ETH/UCY~\cite{pellegrini2010improving, lerner2007crowds} is evaluated leave-one-out with $8$ observed and $12$
predicted frames at $2.5$\,Hz in world coordinates. SDD~\cite{Robicquet2016LearningSE} follows the
TrajNet split with the same $8/12$ horizon and is evaluated in
world-coordinate metres. NBA ~\cite{linou2016nba} uses the SportVU split at
$5$\,Hz with $10$ observed and $20$ predicted frames for $11$ agents ($10$
players and the ball), with coordinates normalised by the court dimensions.

\begin{table}[t]
\centering
\caption{Per-dataset configuration. All other hyperparameters follow the
shared settings above.}
\label{tab:impl_perdataset}
\setlength{\tabcolsep}{6pt}
\scriptsize
\begin{tabular}{lccc}
\toprule
Setting & ETH/UCY & SDD & NBA \\
\midrule
Obs / Pred frames         & 8 / 12 & 8 / 12 & 10 / 20 \\
Latent dim (Eq.~(13))     & 16     & 16     & 24 \\
Neighbours $k$            & 8      & 8      & 10 \\
Distance scale $\sigma_d$ (Eq.~(4)) & 2.0 & 5.0 & 4.0 \\
Dropout                   & 0.1    & 0.1    & 0.15 \\
Epochs                    & 150    & 150    & 200 \\
\bottomrule
\end{tabular}
\end{table}

\section{Directional interaction anisotropy}

We visualize the directional anisotropy induced by the proposed interaction
modeling across multiple representative scenarios.
In our method, interaction strength is modulated using relative bearing, 
motion alignment, and other kinematic descriptors through the anisotropic
kernel defined in Eq.~(10)--(12) of the main paper.
This formulation allows the model to assign different influence to neighbors
depending on their position relative to the target agent.

\begin{figure}[htbp]
\centering
\includegraphics[height=0.21\textheight]{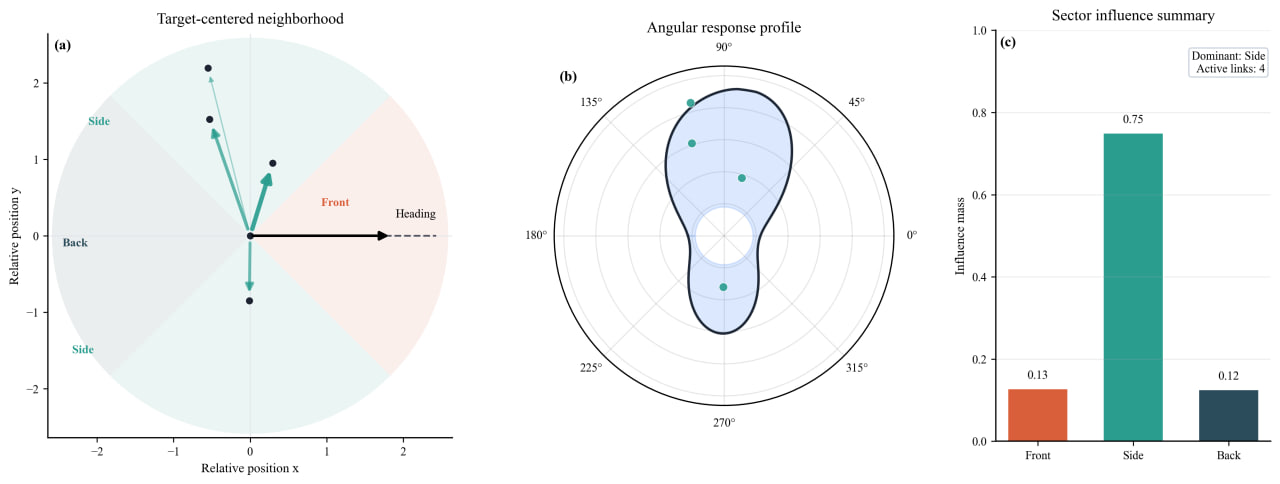}
\includegraphics[height=0.21\textheight]{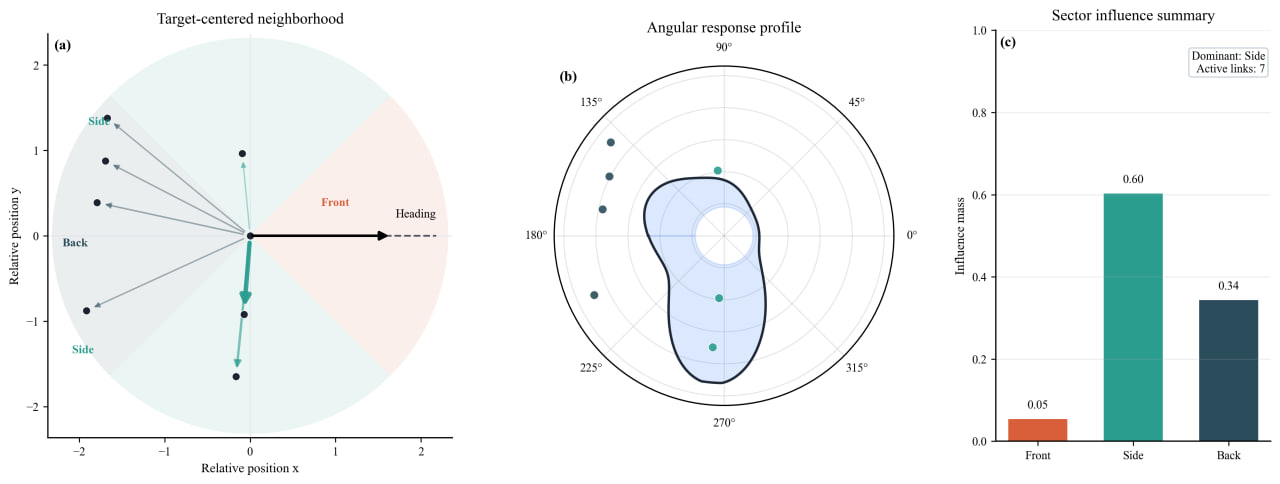}
\includegraphics[height=0.21\textheight]{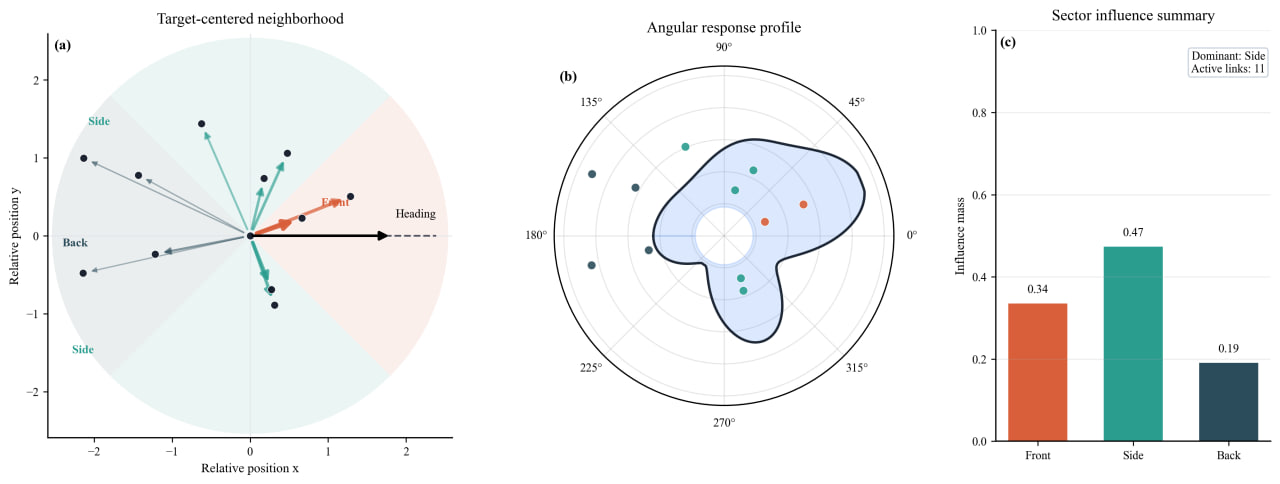}
\includegraphics[height=0.21\textheight]{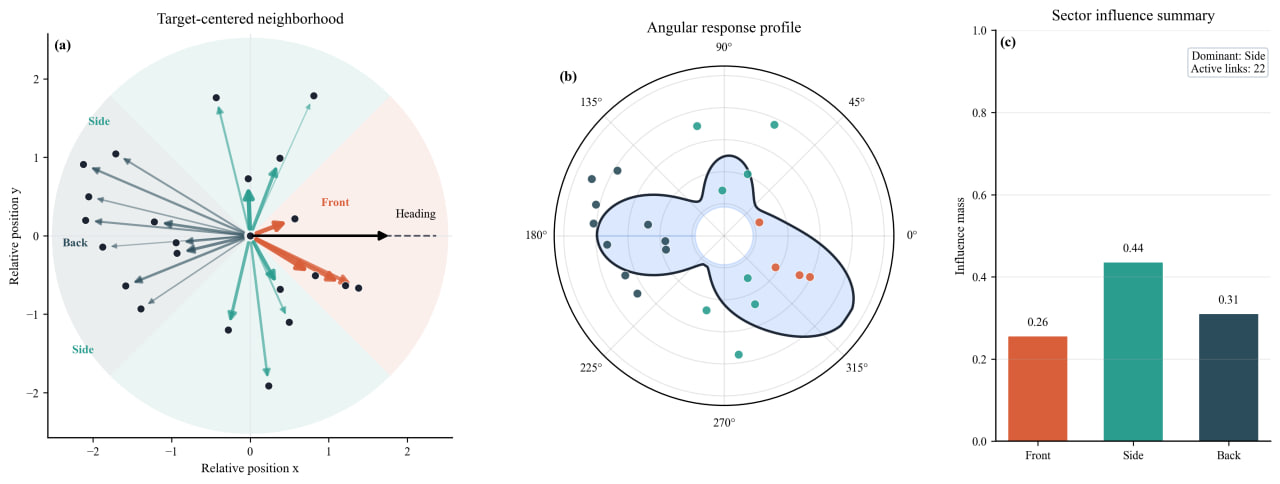}

\caption{ \scriptsize
Directional interaction anisotropy of the proposed model across four representative scenarios. Each row visualizes the directional interaction pattern for a different target pedestrian. (a) Target-centered neighborhood with the target placed at the origin and its heading direction indicated by the black arrow; surrounding space is partitioned into front, side, and back sectors, and arrows denote active interaction links with intensity indicating interaction strength. (b) Angular response profile in polar coordinates, where points correspond to neighboring agents and the blue region indicates accumulated directional interaction strength. (c) Sector-level summary of interaction influence across front, side, and back sectors.
}
\label{fig:anisotropy}
\end{figure}

Figure~\ref{fig:anisotropy} presents several target-centered interaction
neighborhoods, where each row corresponds to a different scene with a 
distinct spatial arrangement of neighboring agents around the target. 
Despite the varying neighborhood geometry, the learned interaction function
consistently produces directional responses aligned with the relative
orientation of neighbors.

For each scenario, the left column shows the spatial neighborhood in the
target-centered coordinate frame, the middle column illustrates the angular
response profile in polar coordinates, and the right column summarizes the
aggregated influence over front, side, and back sectors.
Across all cases, neighbors located in front of the target tend to produce
stronger interaction responses, while agents behind or outside the motion
direction contribute less influence.

Although the proposed interaction function is defined as a smooth directional
kernel, for visualization purposes we partition the surrounding space into
three coarse sectors (front, side, and back) and report the accumulated
interaction strength within each sector.
This discretization is used only for interpretability of the learned
directional response, while the actual model operates with continuous
directional scoring.
Because the kernel is smooth, neighbors located near sector boundaries may
contribute partially to multiple sectors, which explains the presence of
non-zero influence even in directions where no neighbor lies exactly.

The polar response profiles confirm that the interaction kernel is not
isotropic, but concentrates weight in specific angular regions depending on
the scene configuration.
The sector-level summaries further demonstrate that the model adapts its
interaction strength according to relative orientation, indicating that the
anisotropic kernel captures consistent and physically meaningful directional
structure rather than relying on simple distance-based proximity.

In the first scenario, only a few neighbors are present and they are located
mostly in the lateral region of the target-centered frame.
Accordingly, the sector summary shows dominant side influence.
However, small front and back contributions are still observed.
This is expected because the proposed interaction kernel is smooth in the
angular domain, and neighbors positioned near sector boundaries produce
non-zero responses in adjacent directions even when their primary location
is lateral.

In the second scenario, neighbors appear both in the side and rear regions.
The resulting interaction pattern shows stronger responses toward the side,
with particularly large contributions from the two neighbors located on the
right-hand side of the target.
The rear sector also receives noticeable influence, consistent with the
presence of agents behind the target, but the overall response remains
dominated by the lateral configuration.

In the third scenario, several neighbors are positioned in front of the
target, together with additional agents distributed in the lateral region.
The angular response profile therefore expands toward the forward direction,
and the sector summary indicates increased front influence compared to the
previous cases.
At the same time, side influence remains significant due to the presence of
multiple lateral neighbors, demonstrating that the interaction kernel adapts
to the full spatial configuration rather than focusing on a single dominant
direction.

In the fourth scenario, the neighborhood is densely populated around the
target, with agents distributed across front, side, and back regions.
The resulting response profile becomes broader and more complex, reflecting
the combined effect of multiple interaction directions.
Although influence is present in all sectors, the side region still carries
the largest accumulated weight, indicating that the model balances multiple
neighbor contributions while preserving directional sensitivity imposed by
the anisotropic kernel.

\section{Temporal consistency of interaction graphs}

We analyze the temporal behavior of the interaction graph constructed
by the proposed kinematics-conditioned connectivity rule.
In our method, edges are created based on relative distance,
approach rate, motion alignment, and time-to-closest-approach, and are updated
using hysteresis thresholds to avoid rapid changes in connectivity.
This design produces sparse yet temporally stable interaction graphs,
which better reflect physically meaningful relationships between agents.

\begin{figure}[t]
\centering
\includegraphics[width=\linewidth]{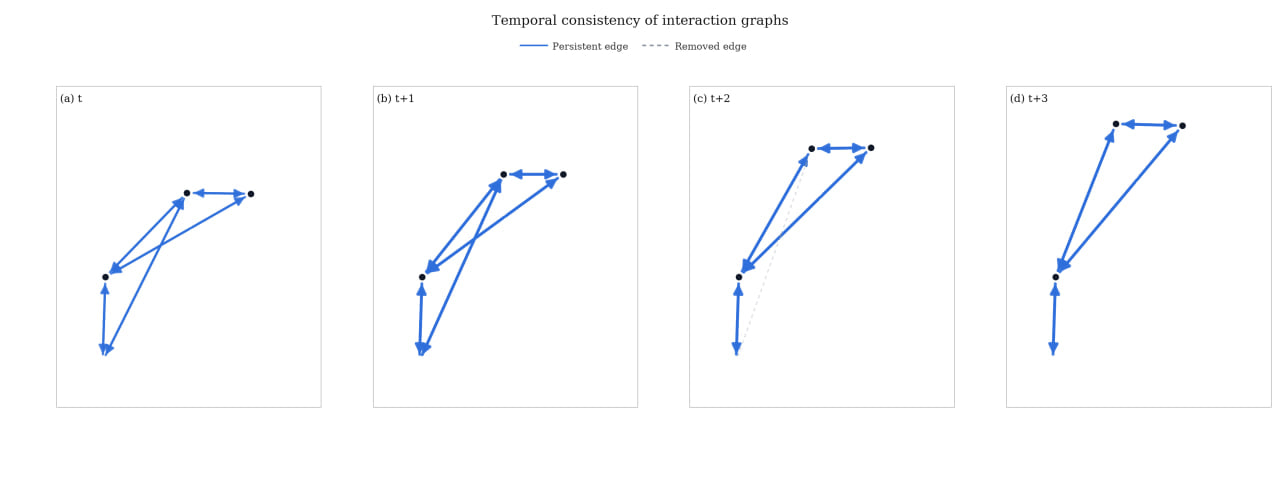}
\caption{
Temporal consistency of the interaction graph.
Edges persist across consecutive timesteps when the kinematic interaction
conditions remain satisfied, while weak interactions are gradually removed.
The hysteresis-based update rule prevents unstable edge switching and
produces temporally smooth connectivity.
}
\label{fig:temporal_graph}
\end{figure}

Figure~\ref{fig:temporal_graph} shows the interaction graph over
consecutive timesteps for a representative scene.
Persistent edges are maintained when agents continue to satisfy the
kinematic conditions, while edges are removed only when the interaction
strength falls below the lower hysteresis threshold.
As a result, the connectivity evolves smoothly over time instead of
flickering between frames.

Across the sequence, interaction links remain stable as relative motion
changes gradually. Kinematically relevant edges persist across frames,
while weaker interactions disappear smoothly as agents move apart or
their motion becomes less relevant. This behavior shows that the proposed
graph construction maintains stable interaction structure while remaining
responsive to motion changes, which is important for trajectory prediction
since unstable connectivity may introduce noise in message passing.

To quantify this behavior, Table~\ref{tab:temporal} reports three
temporal-consistency measures over the observation window, averaged across the
ETH/UCY test sets: the mean edge density, the frame-to-frame edge Jaccard
$|E_t \cap E_{t+1}| / |E_t \cup E_{t+1}|$, and the mean edge lifetime (the
average number of consecutive frames an edge persists). We compare a proximity
graph, MEDIG with the hysteresis replaced by a single fixed threshold, and the
full MEDIG construction. Relative to a single threshold, hysteresis raises the
frame-to-frame Jaccard from $0.89$ to $0.93$ and extends the mean edge lifetime
from $4.7$ to $5.3$ frames, while slightly reducing edge density from $0.31$ to
$0.29$. Thus, the improved temporal stability is not obtained by retaining more
edges. The proximity graph achieves the highest temporal stability
(Jaccard $0.97$), as expected from its dependence on slowly varying inter-agent
distance, but its connectivity is determined purely by geometric proximity.
MEDIG approaches this level of stability while constructing interactions from
kinematic relevance. These measurements confirm quantitatively that the smooth
connectivity in Figure~\ref{fig:temporal_graph} is a property of the construction
rather than of a single scene.

\begin{table}[H]
\centering
\caption{Temporal consistency of the interaction graph over the observation
window, averaged over the ETH/UCY test sets. Edge density, frame-to-frame edge
Jaccard, and mean edge lifetime (in frames; window length $8$).}
\label{tab:temporal}
\setlength{\tabcolsep}{6pt}
\scriptsize
\begin{tabular}{lccc}
\toprule
Graph & Density & Jaccard$(t,t{+}1)$ & Edge lifetime \\
\midrule
Proximity                 & 0.32 & \textbf{0.97} & \textbf{5.7} \\
MEDIG (single threshold)  & 0.31 & 0.89 & 4.7 \\
MEDIG (hysteresis)& \textbf{0.29} & 0.93 & 5.3 \\
\bottomrule
\end{tabular}
\end{table}

\section{Interaction weights vs kinematic cues}

\begin{figure}[H]
\centering
\includegraphics[width=\linewidth]{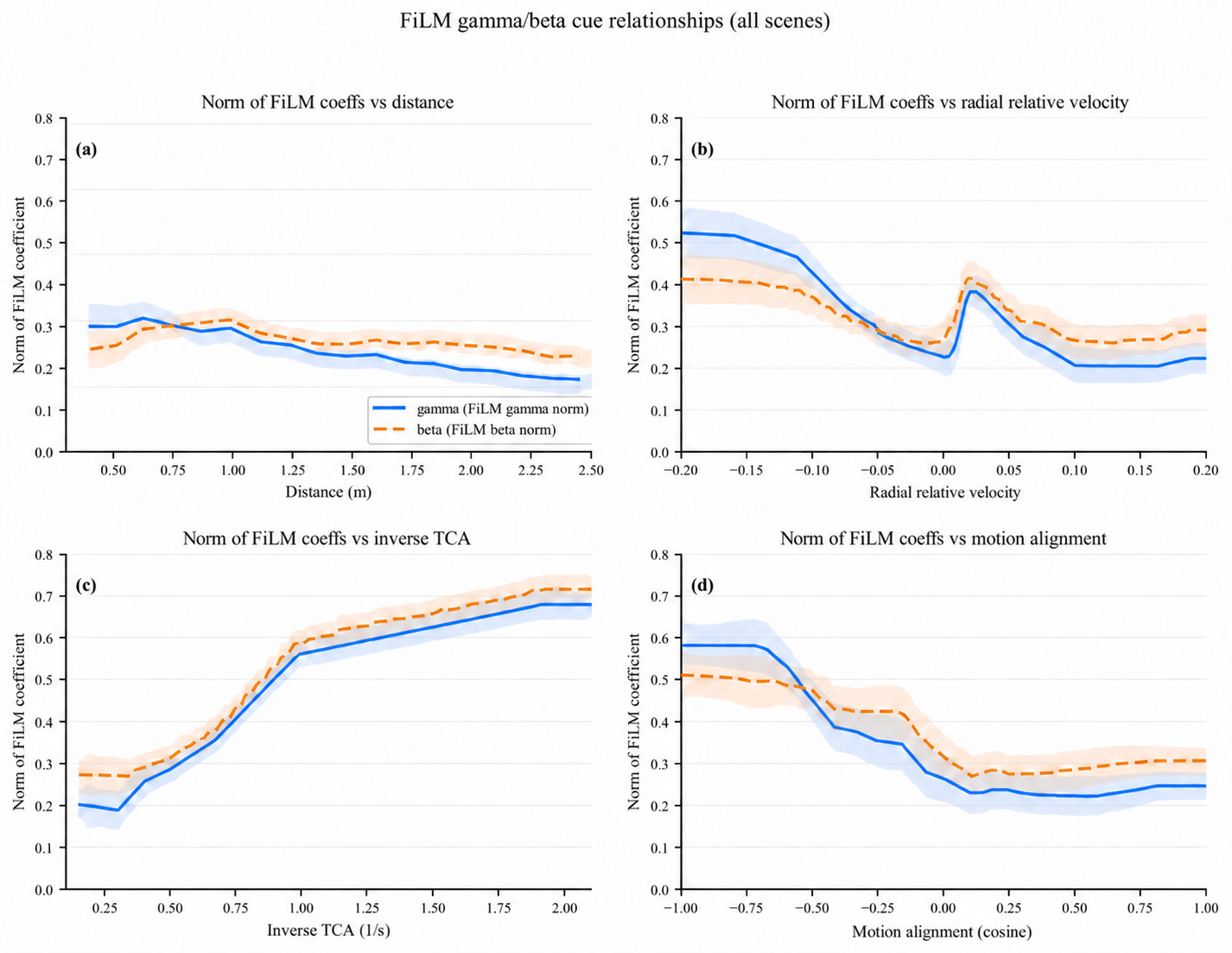}
\caption{
FiLM modulation coefficients as a function of kinematic descriptors.
(a) Coefficients vs distance.
(b) Coefficients vs radial relative velocity.
(c) Coefficients vs inverse time-to-closest-approach.
(d) Coefficients vs motion alignment.
Solid and dotted lines denote the mean normalized FiLM coefficients ($\gamma$ and $\beta$) respectively, across test sets of all five ETH/UCY scenes.
Shaded regions represent the 95$\%$ confidence intervals, shown in blue for $\gamma$ and in orange for $\beta$.
The learned modulation follows physically meaningful trends,
assigning stronger interaction influence to nearby agents,
agents with more imminent closest approach (large inverse TCA),
and agents with compatible motion patterns.
}
\label{fig:weights}
\end{figure}

We analyze how the proposed interaction modeling assigns message-passing
weights as a function of the kinematic descriptors used in the main paper.
Interaction strength depends on relative distance, radial relative velocity,
motion alignment, and time-to-closest-approach, which form the interaction
embedding that modulates graph message passing between agents.
Figure~\ref{fig:weights} shows the FiLM modulation coefficients as a function
of these quantities.
Both FiLM scaling ($\gamma$) and bias ($\beta$) exhibit structured
responses to the kinematic cues.

As distance increases, the modulation strength shows a mild attenuation beyond short range, indicating that nearby agents contribute more strongly to message passing, while the influence of distant agents becomes weaker. This behavior suggests that the model prioritizes local interactions while still allowing moderate influence from mid-range neighbors.
Radial relative velocity exhibits an asymmetric, non-monotonic behavior.
The highest interaction strength occurs for negative radial relative velocities,
corresponding to neighbors moving toward the target agent, which indicates a
higher likelihood of interaction. A smaller local peak appears near zero radial
relative velocity, reflecting situations in which agents move with similar speed
but different directions, such as crossing or perpendicular trajectories, where
interaction is still likely. For positive radial relative velocities, the modulation
strength decreases, consistent with neighbors moving away from the target,
moving in the same direction with higher speed, or remaining behind, all of
which reduce interaction relevance.
The strongest response appears in the inverse time-to-closest-approach descriptor: as inverse TCA increases (i.e., closest approach becomes more imminent), both FiLM coefficients rise sharply, showing that the model assigns larger influence to neighbors involved in more imminent interactions.
Motion alignment shows a clear dependence on the cosine similarity between velocity directions. Interaction strength is highest for negative cosine values, especially near -1, corresponding to agents moving toward each other in opposite directions. As alignment becomes positive, the modulation gradually decreases, indicating weaker interaction when agents move in similar or nearly parallel directions, where immediate interaction relevance is typically lower.
Taken together, these trends show that the four descriptors play complementary roles in modulating interaction strength: distance controls spatial attenuation, inverse TCA captures the imminence of closest approach, radial relative velocity reflects dynamic approach behavior, and motion alignment encodes directional compatibility. This confirms that the learned interaction function follows the intended kinematics-conditioned design and produces physically meaningful message passing.

\section{Additional Analysis on NBA}

\begin{figure}[htpb]
\centering
\includegraphics[width=\linewidth]{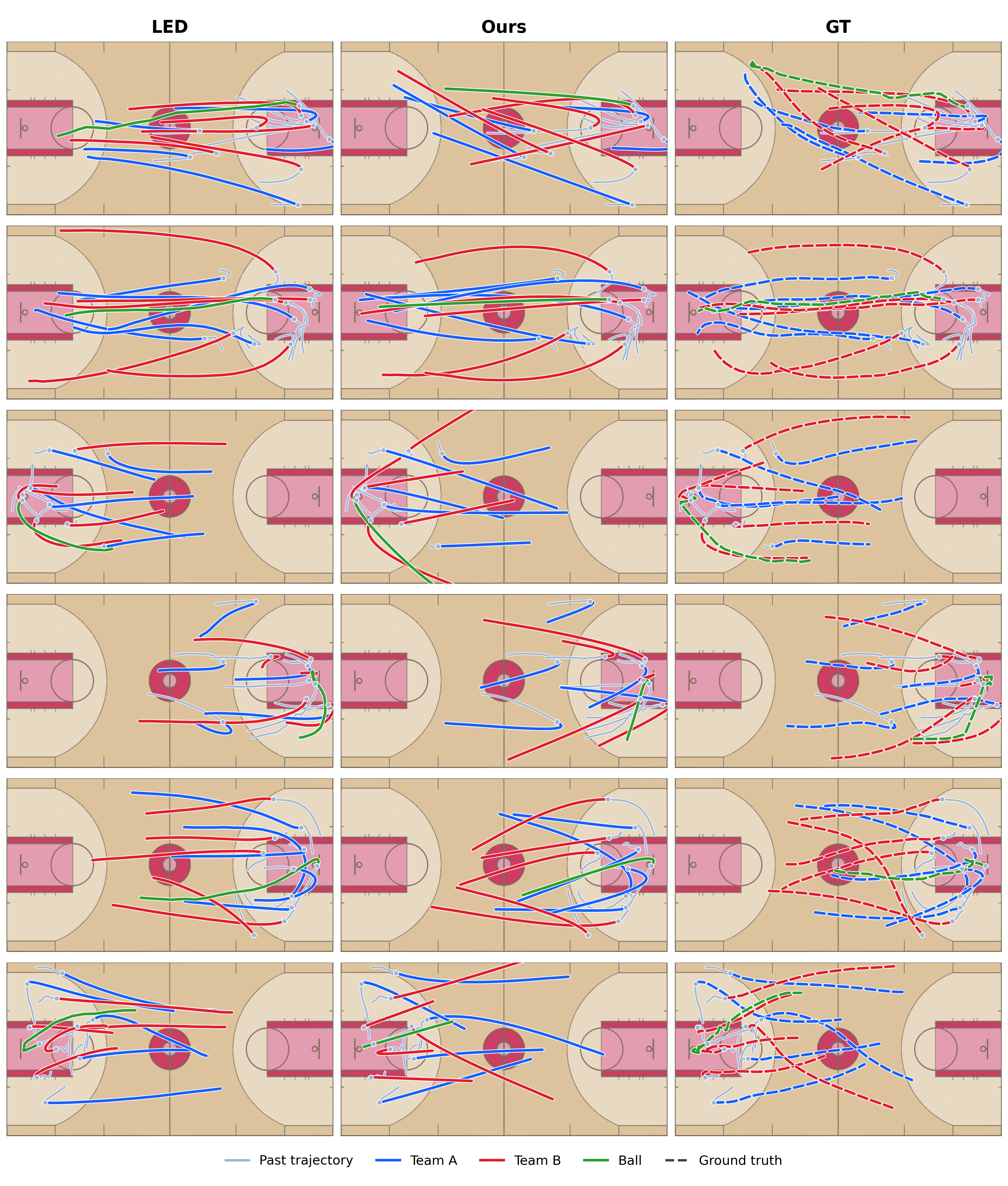}
\caption{
Qualitative comparison on the NBA dataset (best-of-$20$): LED, Ours, and
ground truth (GT). Light: past trajectory; blue and red: the two teams; green:
the ball; dashed: ground-truth future. Representative scenes in which our
predictions align more closely with the ground truth.
}
\label{fig:nba_compare}
\end{figure}

Figure~\ref{fig:nba_compare} compares best-of-$20$ predictions of the proposed
model and LED~\cite{mao2023leapfrog} against the ground truth on the NBA
SportVU dataset. Past trajectories are drawn in light gray, the two teams in
blue and red, and the ball in green; ground-truth futures are dashed. In these
representative scenes, our predictions follow the ground-truth motion more
closely than LED.

\paragraph{Connectivity sensitivity.}
NBA contains $11$ interacting agents, so the top-$k$ limit of $k=10$ is
non-binding when all other agents are potentially available. Sparsity is
instead determined primarily by the kinematic gate, which is applied before
top-$k$ selection. 

Under the NBA configuration in
Table~\ref{tab:impl_perdataset}, MEDIG retains $78.5$ of the $110$ possible
directed edges on average, corresponding to an edge density of $0.714$. Varying the distance scale over $\sigma_d=\{3,4,5\}$ produces edge densities
of $0.69$, $0.71$, and $0.74$, with corresponding $4$\,s minADE/minFDE of
$0.85/1.18$, $0.84/1.16$, and $0.83/1.17$, respectively, indicating that
moderate changes in connectivity density have limited effect on prediction
accuracy. Increasing the activation threshold $\tau_{\mathrm{on}}$ reduces
density to $0.46$ but degrades performance to $0.90/1.25$, suggesting that
overly sparse connectivity removes useful interactions in this setting.

Transferring the ETH/UCY graph configuration directly to NBA gives
$0.92/1.28$ at $4$\,s, compared with $0.84/1.16$ using the NBA-specific
configuration. These results support the use of a denser, dataset-specific
connectivity regime for NBA.

\section{Physical Plausibility}

ADE and FDE measure positional accuracy but do not directly characterize the
dynamics of the predicted trajectories. We therefore additionally evaluate
physical plausibility using the Wasserstein-1 ($W_1$) distance between the
predicted and ground-truth acceleration and jerk distributions on ETH/UCY,
averaged over three random seeds. Acceleration reflects changes in velocity,
while jerk, the temporal derivative of acceleration, is particularly sensitive
to abrupt or dynamically irregular motion. Lower $W_1$ indicates that the
predicted dynamical distributions more closely match those of the
ground-truth trajectories.

Table~\ref{tab:physical_plausibility} compares the full model with variants
without kinematic conditioning and without the smoothness objective. The full
model achieves the lowest distance for both acceleration and jerk,
with $W_1(a)=0.228$ and $W_1(j)=0.996$. Removing kinematic conditioning
increases these values to $0.297$ and $1.311$, respectively, despite retaining
the smoothness loss. This shows that matching realistic motion dynamics is not
explained by smoothness regularization alone: the kinematics-conditioned
interaction mechanism also contributes to producing trajectories whose local
dynamics better resemble the ground truth.

Removing $L_{\mathrm{smooth}}$ causes the largest degradation, increasing
$W_1(a)$ to $0.540$ and $W_1(j)$ to $2.477$. Together, these results indicate
that the two components play complementary roles. The smoothness objective
discourages abrupt temporal variation, while kinematic conditioning helps the
model account for interaction-driven changes in motion. Combined with the
ADE/FDE results in the main paper, these findings indicate that the full model
achieves accurate trajectory prediction while more closely matching the
ground-truth dynamical distributions.

\begin{table}[h]
\centering
\caption{\textbf{Physical plausibility on ETH/UCY.} Wasserstein-1 distance
between predicted and ground-truth acceleration and jerk distributions.}
\label{tab:physical_plausibility}
\scriptsize
\setlength{\tabcolsep}{6pt}
\begin{tabular}{lcc}
\toprule
Method & $W_1(a)$ [m/s$^2$] $\downarrow$ & $W_1(j)$ [m/s$^3$] $\downarrow$ \\
\midrule
w/o Kinematics & 0.297 & 1.311 \\
w/o $L_{\mathrm{smooth}}$ & 0.540 & 2.477 \\
Full & \textbf{0.228} & \textbf{0.996} \\
\bottomrule
\end{tabular}
\end{table}

\bibliographystyle{splncs04}
\bibliography{main}

%% file: sec/0_abstract.tex
\begin{abstract}

Accurate human trajectory prediction requires modeling not only where pedestrians are, but also how they are moving relative to one another. Existing graph-based methods often rely on geometric proximity or learned attention, which can produce interaction structures weakly tied to physical motion cues. We propose a kinematics-conditioned graph hypernetwork for trajectory prediction, where inter-agent messages are generated from relative motion features such as velocity, alignment, directionality, and time-to-closest-approach. This allows the model to capture anisotropic and context-dependent interactions while maintaining sparse, motion-grounded connectivity. The proposed framework supports both deterministic and probabilistic prediction and reduces unnecessary message passing without sacrificing accuracy. Experiments on widely used trajectory prediction benchmarks show that our approach improves over representative graph-based baselines and remains competitive with stronger recent attention- and transformer-style models, while offering stable sparsification and lower computational overhead. Code will be made publicly available.

\keywords{Trajectory Prediction \and Graph Neural Networks \and Anisotropic Message Passing \and Interaction Modeling}

\end{abstract}

%% file: sec/1_introduction.tex
\section{Introduction}
\label{sec:intro}
Modeling pedestrian motion in shared spaces remains a long-standing challenge in computer vision. Accurate trajectory prediction enables safer autonomous navigation, efficient crowd simulation, and socially compliant robot planning~\cite{rudenko2020human, uhlemann2024evaluating, Robicquet2016LearningSE}. While recent models achieve remarkable predictive accuracy, they often overlook the \emph{kinematic structure} governing real human motion. In particular, they fail to account for how velocity, heading, and physical feasibility shape social interactions. Effective trajectory prediction requires reasoning about these underlying motion dynamics rather than solely on data-driven learned relations.


Graph-based trajectory predictors capture interactions through message passing \cite{kosaraju2019social, mohamed2020social, salzmann2020trajectron++}, while sequence and transformer-based approaches treat trajectories as sequences of tokens, modeling relations through attention \cite{giuliari2021transformer, yuan2021agentformer, chen2022fully, shi2023trajectory, xu2023eqmotion, zhou2022spatiotemporal}.
Despite architectural differences, most existing approaches define interactions via fully connected relations.
Fully connected graphs are computationally inefficient, and they produce noisy, redundant messages that weaken meaningful interactions. By ignoring directional dependencies and physical feasibility, they generate excessive connections that are inconsistent with the naturally sparse structure of real human interactions. This noise and inefficiency hinder scalability and accurate modeling of motion dynamics \cite{mohamed2020social, mendieta2021carpe}.

We instead construct interaction graphs from measurable pairwise motion cues: relative velocity, heading alignment, distance attenuation, time-to-closest-approach (TCA), and bearing. These cues indicate whether two agents are approaching, moving together, or likely to occupy nearby space in the future. Prior empirical analysis has shown that such quantities correlate with trajectory adjustments in pedestrian datasets~\cite{amirian2020opentraj}.

We propose a \emph{kinematics-conditioned graph convolution network} that performs direction-aware message passing on a sparse interaction graph. 
Edges are gated by measurable motion cues so that information flow reflects human motion dynamics. 
A lightweight hypernetwork \cite{ha2016hypernetworks} converts these cues into FiLM-style \cite{perez2018film, brockschmidt2020gnn} modulation parameters that enhance anisotropic message propagation.

Finally, the framework includes two decoder variants used in distinct setups: a deterministic decoder for single-mode trajectory prediction and a probabilistic CVAE-based decoder for multimodal generation \cite{NIPS2015_8d55a249}.
We evaluate the proposed model on the ETH/UCY, SDD, and NBA benchmarks, where it achieves competitive accuracy with substantially lower computational cost.

Our contributions can be summarized as:
\begin{itemize}

    \item A \textbf{kinematics-based sparse graph construction} strategy that builds directed interaction edges using kinematic features, which leads to more efficient information propagation for learning social interactions.   
    \item A \textbf{kinematics-conditioned graph hypernetwork} that generates edge-specific FiLM modulation parameters from pairwise motion cues and global scene context.
    \item A \textbf{direction-aware anisotropic message-passing module} that blends front, side, and back interaction filters using Gaussian angular kernels, enabling continuous orientation-dependent edge weighting.
    \item Extensive evaluation on ETH/UCY, SDD, and NBA benchmarks, showing improved accuracy and efficiency.

\end{itemize}

%% file: sec/2_related.tex
\section{Related work}

\textbf{Early Trajectory Forecasting.}
Early trajectory forecasting methods grounded motion in explicit dynamical or probabilistic rules, social-force formulations~\cite{helbing1995social, 5459260}, and Gaussian-process motion priors~\cite{wang2005gaussian}.
These models provided interpretable physical structure but lacked the flexibility to capture complex social interactions, diverse behaviors, and scene-dependent variability. With the emergence of deep learning, recurrent architectures began modeling motion directly from data \cite{alahi2016social, gupta2018social, kothari2021human, sadeghian2019sophie, zhang2019sr}, improving predictive accuracy but still encoding interactions implicitly within pooled hidden states, without an explicit relational representation.

\textbf{Graph-based Trajectory Prediction.}
Graph-structured models make interactions explicit by representing agents as nodes connected through learned relations.
Proximity-based methods such as Social-STGCNN \cite{mohamed2020social} use distance-defined edges, while attention-driven models including STGAT \cite{huang2019stgat} and Social-BiGAT \cite{kosaraju2019social} infer interaction strengths from data. Probabilistic models such as Trajectron++ \cite{salzmann2020trajectron++} incorporate graph structure within a CVAE to capture multimodal futures, and recent variants introduce sparsity or directionality, e.g., SGCN \cite{shi2021sgcn} prunes redundant edges and T-GNN \cite{xu2022adaptive} learns transferable interaction graphs for cross-domain generalization.

Despite these differences, existing graph-based predictors typically define edges or weights through geometric proximity or unconstrained attention. Although velocity or heading may be included as node features, the interaction function itself is not explicitly parameterized by physically meaningful descriptors such as relative velocity, alignment, or time-to-collision. Thus, these models expose relational structure but do not directly encode whether one agent is kinematically relevant or constraining another’s motion.

\textbf{Sequence-Based Models.}
Recurrent models such as Social-LSTM \cite{alahi2016social} and Social-GAN \cite{gupta2018social} capture temporal dependencies through pooled hidden states, but represent interactions implicitly and struggle with long-range dependencies.
Transformer architectures \cite{vaswani2017attention}, including AgentFormer \cite{yuan2021agentformer} and TUTR \cite{shi2023trajectory}, improve temporal reasoning by attending over tokenized motion histories.
However, their interaction modeling remains data-driven rather than physically grounded: attention weights capture statistical correlations, offer limited interpretability \cite{kothari2021interpretable}, scale quadratically with the number of agents and timesteps, and provide no explicit mechanism for directionality or anisotropic influence.
These limitations motivate models that retain expressive temporal reasoning while incorporating explicit physical structure.

\textbf{Physically-Aware and Directional Modeling.}
Several works revisit physical cues for interaction modeling, drawing inspiration from classical social-force formulations where collision avoidance and directional pressure are explicitly encoded \cite{helbing1995social}. More recent approaches model anisotropy through handcrafted angular sectors, such as front/side/rear partitions in SocialCircle \cite{wong2024socialcircle}, highlighting the importance of orientation in human motion. However, these schemes rely on rigid, predefined ranges that cannot express continuous or context-dependent interaction strengths.

Empirical analyses such as OpenTraj~\cite{amirian2020opentraj} further show that measurable kinematic indicators, including time-to-collision, relative velocity, and heading alignment, correlate strongly with interaction intensity and trajectory adjustments. To our knowledge, prior graph-based trajectory predictors have not directly parameterized their interaction function in terms of such continuous, physically meaningful descriptors; when these quantities appear, they are typically used as auxiliary features rather than as the primary conditioning signals that govern edge strengths.

%% file: sec/3_method.tex
\section{Method}
Given the observed trajectories of $N$ pedestrians over $T_\text{obs}$ timesteps, $ \mathbf{X}=\{\mathbf{x_i^t}\in\mathbb{R}^2\,|\, i=1,\dots,N,\ t=1,\dots,T_\text{obs}\}$, the goal is to predict their future positions for the next $T_\text{pred}$ frames, $\hat{\mathbf{Y}}=\{\mathbf{\hat{y}_i^{t}\}}_{t=T_\text{obs}+1}^{T_\text{obs}+T_\text{pred}}$. The challenge lies in anticipating how each agent’s motion evolves under \emph{social} and \emph{physical} interactions that unfold dynamically in space and time. We formulate this as \textbf{direction-aware message passing} on a kinematics-conditioned graph, where edges encode interpretable kinematic dependencies rather than arbitrary proximity or attention weights.

\subsection{Model Overview}
\begin{figure*}[t]
    \centering
    \includegraphics[width=\textwidth]{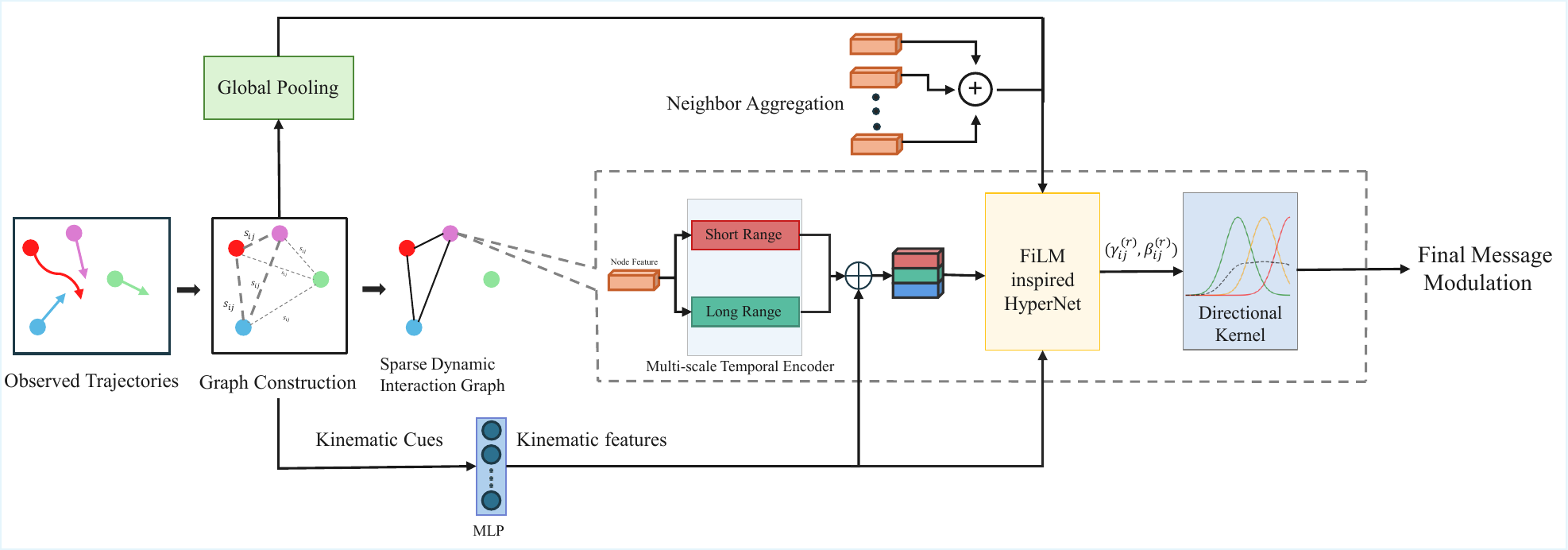}
    \caption{
    Overview of our framework. 
    The model constructs a kinematics-conditioned interaction graph from observed trajectories, extracts multi-scale temporal features, and performs direction-aware message passing using a FiLM-inspired hypernetwork modulated by continuous angular kernels. The resulting anisotropic messages are aggregated to produce physically consistent future trajectory predictions.
    }
    \label{fig:method}
\end{figure*}

Our framework integrates four complementary components:  
(1) a \textbf{kinematics\-conditioned interaction graph} that captures kinematic dependencies among agents using measurable motion quantities,  
(2) a \textbf{multi-scale temporal encoder} that summarizes short- and long-term motion cues for each pedestrian,  
(3) a \textbf{direction-aware message passing} module that modulates interaction strength using FiLM-generated anisotropic filters conditioned on kinematic features, and  
(4) a \textbf{decoding module} that predicts physically consistent future trajectories, implemented in two variants: a deterministic decoder for single-mode prediction and a probabilistic CVAE-based decoder for multimodal trajectory generation.  
These modules form a fully differentiable, unified architecture that couples interpretable physical reasoning with adaptive relational learning. See
Fig.~\ref{fig:method} for an overview of the full model.

\subsection{Kinematics-Conditioned Interaction Graph (MEDIG)}
We model each scene as a \textbf{Momentum-Enhanced Dynamic Interaction Graph} $\mathcal{G}=(\mathcal{V},\mathcal{E})$, where each node $i\!\in\!\mathcal{V}$ corresponds to a pedestrian, and directed edges $(i,j)\!\in\!\mathcal{E}$ represent potential influence from agent $j$ to $i$. 
Most existing graph-based models construct either fully connected interaction graphs or learn sparse structures through data-driven attention, which often capture correlations rather than kinematically grounded relations. 
In contrast, MEDIG defines edge affinities explicitly from measurable kinematic quantities and produces interpretable, direction-aware, and physically consistent connectivity.

Let $\mathbf{x}_i^t\!\in\!\mathbb{R}^2$ denote the position of agent $i$ at time $t$, and $\mathbf{v}_i^t = \mathbf{x}_i^t - \mathbf{x}_i^{t-1}$ its velocity. For every pair $(i,j)$, we derive measurable \emph{kinematic descriptors} from the relative displacement $\mathbf{d}_{ij}=\mathbf{x}_j-\mathbf{x}_i$ and relative velocity $\mathbf{v}_{ij}=\mathbf{v}_j-\mathbf{v}_i$, capturing how $j$’s motion affects $i$ under physical constraints. 

We compute four trajectory-derived descriptors \cite{amirian2020opentraj}: an \emph{approach score} $s_{ij}^{\text{app}}$, an \emph{inverse time-to-closest-approach score} $s_{ij}^{\text{tca}}$, a \emph{motion alignment score} $s_{ij}^{\text{align}}$, and a \emph{distance attenuation score} $s_{ij}^{\text{dist}}$. Together, these quantities capture complementary physical aspects of pairwise interactions, including whether two agents are moving toward each other, how soon their relative motion reaches closest approach, whether their motion directions are aligned, and how their influence decays with spatial distance.

\begin{align}
\hat{\mathbf{d}}_{ij}
&=
\frac{\mathbf{d}_{ij}}
{\|\mathbf{d}_{ij}\|+\epsilon},
\qquad
s^{\text{app}}_{ij}
=
-\hat{\mathbf{d}}_{ij}\!\cdot\!\mathbf{v}_{ij},
\label{eq:app}\\
\mathrm{TCA}_{ij}
&=
-\frac{\mathbf{d}_{ij}\!\cdot\!\mathbf{v}_{ij}}
{\|\mathbf{v}_{ij}\|^2+\epsilon},
\qquad
s^{\text{tca}}_{ij}
=
\frac{1}{\mathrm{TCA}_{ij}+\epsilon},
\label{eq:tca}\\
s^{\text{align}}_{ij}
&=
\cos(\tilde{\mathbf{v}}_i,\tilde{\mathbf{v}}_j),
\qquad
\tilde{\mathbf{v}}_i
=
\alpha\,\mathbf{v}_i^{t}
+
(1-\alpha)\mathbf{v}_i^{t-1},
\label{eq:align}\\
s^{\text{dist}}_{ij}
&=
\exp(-\|\mathbf{d}_{ij}\|^2/\sigma_d^2).
\label{eq:dist}
\end{align}

Equations~\ref{eq:app}--\ref{eq:dist} respectively capture the approach rate between agents, the inverse time-to-closest-approach, the smoothed-direction alignment of their motion vectors, and a Gaussian distance attenuation term that ensures local sensitivity. For numerical stability, $s^{\text{tca}}_{ij}$ is clamped to $[-10,10]$; negative TCA values are retained and therefore yield negative interaction scores. Furthermore, a relative bearing angle $b_{ij} = \angle(\mathbf{d}_{ij},\,\tilde{\mathbf{v}}_i)$ quantifies the orientation of $j$ with respect to $i$'s motion, indicating whether $j$ lies in front, beside, or behind $i$. The overall \emph{interaction potential} combines these cues: 
\begin{equation} 
\psi_{ij} = \lambda_1 s^{\text{app}}_{ij} + \lambda_2 s^{\text{tca}}_{ij} + \lambda_3 s^{\text{align}}_{ij} + \lambda_4 s^{\text{dist}}_{ij} + \lambda_5 \max(\cos b_{ij},0),
\end{equation} 

where $\lambda_1,\dots,\lambda_5$ are tunable coefficients chosen empirically. 
High interaction scores $\psi_{ij}$ indicate that agent $j$ is likely to influence the future motion of agent $i$ under current kinematic conditions. 
Edges are activated when $\psi_{ij}\!\geq\!\tau_\text{on}$ and remain active while $\psi_{ij}\!\geq\!\tau_\text{off}$, forming a hysteretic, temporally coherent topology that suppresses frame-to-frame fluctuations. 
This formulation produces a \emph{sparse, interaction graph}, where edges encode potential motion influence rather than simple spatial proximity. 
To preserve stable sparsity, we retain the top-$k$ strongest interactions
and use exponential momentum smoothing to stabilize connectivity across frames~\cite{shi2021sgcn}.

Each active edge retains its normalized vector
$\mathbf{s}_{ij}=[s^{\text{app}}_{ij}, s^{\text{tca}}_{ij}, s^{\text{align}}_{ij}, s^{\text{dist}}_{ij}]$
and relative bearing $b_{ij}$.
This design produces a graph whose structure emerges directly from measurable motion cues,
allowing the model to reason about directional interaction structure in a physically interpretable manner
rather than relying on arbitrary proximity or attention weights.

\subsection{Temporal Node Representation}

Having established a sparse, kinematics-conditioned interaction graph,
we next construct node embeddings that capture each pedestrian’s temporal motion context.
While the graph encodes inter-agent dependencies, effective message passing also requires
robust representations of individual dynamics over time.
To this end, we process the observed displacement sequence of each agent
through a dual-branch temporal encoder designed to capture both short-term fluctuations
and longer-range motion trends.

Two learnable 1-D convolutional paths with kernel sizes $3$ and $5$ operate on the displacement history,
extracting fine-grained and broader temporal features, respectively.
In parallel, we derive compact kinematic statistics,
\[
\mathbf{k}_i = [\,\mathbf{v}_i^{t},\ \bar{\mathbf{v}}_i,\ \mathbf{a}_i^{t}\,],
\]
representing instantaneous velocity, mean velocity, and acceleration,
computed from the same displacement sequence.
Concatenating these with the convolutional outputs,
\[
\mathbf{z}_i = [\,f_i^{(3)},\,f_i^{(5)},\,\mathbf{k}_i\,],
\]
yields a motion embedding that fuses short- and longer-range temporal cues
with interpretable kinematic quantities.
This physically grounded embedding serves as the node initialization
for direction-aware message passing in subsequent layers.

\subsection{Direction-Aware Message Passing}

Having obtained richer node features after temporal encoding,
we now describe how information is propagated across agents through
\textbf{FiLM-modulated message passing}.
The objective is to enable each pedestrian’s representation
to integrate information from its neighbors adaptively,
guided by their relative kinematics and directional context.

In a standard graph convolutional layer, each node aggregates
uniformly transformed messages from its neighbors:
\begin{equation}    
\mathbf{h}_i^{(l+1)} =
\sigma\!\Big(
\sum_{j \in \mathcal{N}(i)}
W^{(l)} \mathbf{h}_j^{(l)}
\Big),
\end{equation}

where $\mathbf{h}_i^{(l)}$ denotes the feature representation of node $i$
at layer $l$, $W^{(l)}$ is a learnable transformation,
and $\sigma$ is a nonlinear activation.
Such uniform aggregation ignores physical constraints: all neighbors contribute equally, regardless of their
relative motion or spatial configuration.

To achieve adaptive and context-dependent interactions,
we introduce a \emph{FiLM-inspired hypernetwork}
that predicts layer-specific per-edge modulation parameters
$(\gamma_{ij}^{(l)},\beta_{ij}^{(l)})$.
The message $\mathbf{m}_{ij}^{(l)}$ from node $j$ to node $i$ is formulated as
\begin{equation}    
\mathbf{m}_{ij}^{(l)} =
\gamma_{ij}^{(l)} \odot \mathbf{h}_j^{(l)} + \beta_{ij}^{(l)},
\end{equation}

where $\odot$ denotes element-wise multiplication.
Each message is therefore conditioned on edge-specific modulation,
allowing the network to adjust influence strength dynamically.
The updated node features are obtained as
\begin{equation}    
\mathbf{h}_i^{(l+1)} =
\sigma\!\Big(
U^{(l)}
\sum_{j \in \mathcal{N}(i)}
\mathbf{m}_{ij}^{(l)}
\Big),
\end{equation}

with $U^{(l)}$ denoting a shared post-aggregation projection.

Unlike conventional GNN-FiLMs that generate $(\gamma_{ij}, \beta_{ij})$
solely from node embeddings, our model conditions them on
interpretable physical quantities.
A shared hypernetwork $g_\phi$ predicts the modulation parameters
using the normalized kinematics descriptor $\mathbf{s}_{ij}$,
the fused node-pair context $[\mathbf{h}_i, \mathbf{h}_j]$,
and a global scene gate $g_{\text{scene}}$ obtained by mean-pooling the node features followed by a two-layer MLP, that encodes holistic crowd dynamics.
While the graph remains sparsified through kinematics-based connectivity,
this global context pathway ensures that each interaction remains aware
of the overall motion flow and density of the scene,
bridging local physical relations with global situational awareness:
\begin{equation}
(\gamma_{ij}^{(l,r)},\beta_{ij}^{(l,r)}) =
g_\phi^{(r)}\!\big([\mathbf{s}_{ij},\mathbf{h}_i^{(l)},\mathbf{h}_j^{(l)},g_{\text{scene}}]\big).
\label{eq:hypernet}
\end{equation}
for $r \in \{\text{front},\text{side},\text{back\}}$ where $r$ represents the relative direction. Each head $g_\phi^{(r)}$ corresponds to a canonical spatial role.

Directional selectivity is modeled by blending these role-specific parameters
through Gaussian angular kernels centered at
front, side, and back orientations:
\begin{equation}
w_r(b_{ij}) =
\frac{\exp(-(\Delta b_{ij,r})^2/2\sigma_{\text{ang}}^2)}
{\sum_{r'}\exp(-(\Delta b_{ij,r'})^2/2\sigma_{\text{ang}}^2)}.
\label{eq:angular_kernel}
\end{equation}
Here, $\Delta b_{ij,r}$ is the angular difference from role $r$; we fix
$\sigma_{\text{ang}}=0.6$ rad across datasets.
The final modulation is obtained as

\begin{equation}
(\gamma_{ij}^{(l)},\beta_{ij}^{(l)})
= \sum_r w_r(b_{ij})
(\gamma_{ij}^{(l,r)},\beta_{ij}^{(l,r)}).
\end{equation}

yielding a smooth, continuously varying anisotropy
that biases message flow toward physically plausible directions
without discrete role assignments.

Combining these definitions,
the final message that will be used in node feature updates can be expressed as
\begin{equation}
\mathbf{m}_{ij}^{(l)} =
\sum_r w_r(b_{ij})
\big(\gamma_{ij}^{(l,r)}\odot\mathbf{h}_j^{(l)}
+\beta_{ij}^{(l,r)}\big).
\end{equation}

Residual connections and layer normalization follow each layer.
Stacking multiple such blocks extends the spatio-temporal receptive field,
enabling higher-order reasoning about collective motion and interaction dynamics.

\subsection{Probabilistic Trajectory Decoder}

Future positions are generated through a conditional variational decoder designed to capture multimodal intent. Given the refined node feature $\mathbf{h}_i$, we first map it to a latent distribution via a prior network, yielding $(\boldsymbol{\mu}_i,\boldsymbol{\sigma}_i^2)=\text{Prior}(\mathbf{h}_i)$. To represent distinct potential future behaviors, we draw $K$ latent samples $\mathbf{z}_{i,k}\sim\mathcal{N}(\boldsymbol{\mu}_i,\boldsymbol{\sigma}_i^2)$.

Each sampled latent vector is then combined with the original node feature to generate a distinct trajectory hypothesis:

\begin{equation}
    \hat{\mathbf{Y}}_{i,k} =
f_\theta(\mathbf{h}_i + W_z\mathbf{z}_{i,k}),
\quad k=1,\dots,K.
\end{equation}
Here, $f_\theta$ denotes the trajectory generator with weights $\theta$, and $W_z$ is a learnable projection matrix that aligns the latent samples with the dimensional space of the node features.
During training, a posterior network conditions on the ground truth; at inference, the \emph{best-of-$K$} sample represents the model’s prediction.

\subsection{Training Objective}

Our model is trained end-to-end using five complementary loss terms that encourage accurate reconstruction, structured uncertainty, and physically consistent motion. Specifically, we combine a reconstruction loss defined in Eqs.~\eqref{eq:recon_det} and \eqref{eq:recon_div}, which ensures trajectory fidelity and diversity, a KL regularization term defined in Eq.~\eqref{eq:kl} that enforces a smooth latent space, and three kinematics-inspired auxiliary losses for endpoint accuracy, directional coherence, and temporal smoothness, defined in Eqs.~\eqref{eq:endp}, \eqref{eq:dir}, and \eqref{eq:smooth}, respectively.

We employ two training setups: a deterministic variant for single-mode prediction and a diverse variant for multimodal generation.  
For the deterministic setup, a simple Mean Squared Error (MSE) loss measures pointwise deviation between the predicted trajectory $\hat{\mathbf{y}}$ and the ground truth $\mathbf{y}$:
\begin{equation}    
\mathcal{L}_\text{recon}^{\text{det}} = 
\| \hat{\mathbf{y}} - \mathbf{y} \|_2^2.
\label{eq:recon_det}
\end{equation}

In the diverse setup, we adopt a Best-of-$K$ Smooth-$L_1$ loss which selects the trajectory with the smallest error among the $K$ stochastic samples:
\begin{equation}    
\mathcal{L}_\text{recon}^{\text{div}} =
\min_{k \in \{1,\ldots,K\}}
\| \hat{\mathbf{y}}_k - \mathbf{y} \|_{1,\text{smooth}},
\label{eq:recon_div}
\end{equation}


Within the conditional variational autoencoder framework, we regularize the latent variable $\mathbf{z}$ by minimizing

\begin{equation}    
\mathcal{L}_\text{KL} =
D_\text{KL}\!\big(q_\phi(\mathbf{z}|\mathbf{x},\mathbf{y})\,\|\,p_\theta(\mathbf{z}|\mathbf{x})\big),
\label{eq:kl}
\end{equation}
which aligns the posterior and prior distributions to promote a well-structured latent space. The coefficient $\beta_\text{KL}$ is gradually annealed during training to balance reconstruction fidelity and latent diversity.
We further impose three auxiliary losses to ensure physically plausible motion dynamics:

  \begin{equation}      
\mathcal{L}_\text{endp}
= \big\|\hat{\mathbf{y}}_{T_\text{pred}} - \mathbf{y}_{T_\text{pred}}\big\|_2^2,
\label{eq:endp}
  \end{equation}

  \begin{equation}
\mathcal{L}_\text{dir}
= 1 -
\frac{
\langle \hat{\mathbf{v}}_{T_\text{pred}},\, \mathbf{v}_{T_\text{pred}}\rangle}{
\|\hat{\mathbf{v}}_{T_\text{pred}}\|_2
\,\|\mathbf{v}_{T_\text{pred}}\|_2+\varepsilon},
\label{eq:dir}
\end{equation}

  \begin{equation}
\mathcal{L}_\text{smooth}
= \frac{1}{T_\text{pred}-1}
\sum_{t=2}^{T_\text{pred}}
\big\| \hat{\mathbf{v}}_t - \hat{\mathbf{v}}_{t-1} \big\|_2^2,
\label{eq:smooth}
\end{equation}

Specifically, $\mathcal{L}_\text{endp}$ penalizes endpoint deviations to emphasize destination accuracy. $\mathcal{L}_\text{dir}$ enforces alignment between the predicted and ground-truth final velocity directions. Finally, $\mathcal{L}_\text{smooth}$ regularizes second-order temporal derivatives, discouraging abrupt acceleration changes and promoting smooth trajectories. Together, these losses encourage realistic and kinematically consistent motion across time. All components are optimized jointly with the composite objective:
\[
\mathcal{L}
= \mathcal{L}_\text{recon}
+ \beta_\text{KL}\,\mathcal{L}_\text{KL}
+ \lambda_\text{endp}\,\mathcal{L}_\text{endp}
+ \lambda_\text{dir}\,\mathcal{L}_\text{dir}
+ \lambda_\text{smooth}\,\mathcal{L}_\text{smooth}.
\]
This formulation ensures that predicted trajectories are simultaneously diverse, accurate, and physically consistent with real-world motion patterns.

%% file: sec/4_experiments.tex
\section{Experiments}

\textbf{Experimental Setup.}
We evaluate our model on three widely used trajectory benchmarks:
\textbf{ETH/UCY}~\cite{pellegrini2010improving, lerner2007crowds},
the \textbf{Stanford Drone Dataset (SDD)}~\cite{Robicquet2016LearningSE},
and the \textbf{NBA} SportVU dataset~\cite{linou2016nba}.
ETH/UCY contains five scenes (\textit{ETH}, \textit{HOTEL},
\textit{UNIV}, \textit{ZARA1}, \textit{ZARA2}) of pedestrian interactions,
evaluated leave-one-out, while SDD provides top-view aerial videos of
crowded scenes with complex multi-agent dynamics. For both, trajectories
are sampled at $2.5$\,Hz with $8$ observed ($3.2$\,s) and $12$ predicted
($4.8$\,s) frames, following the standard protocol. NBA captures interactive basketball play among $11$ agents ($10$ players and the ball);
following common practice, we observe $2.0$\,s ($10$ frames) and predict
$4.0$\,s ($20$ frames) at $5$\,Hz. We report Average and Final
Displacement Error (ADE/FDE) for deterministic prediction, and
minADE/FDE~\cite{gupta2018social} across $20$ samples for probabilistic
prediction. To assess interpretability and efficiency, we additionally
measure the average number of active edges per scene and the computational
cost in FLOPs and runtime per forward pass.
All experiments are implemented in PyTorch and trained on an NVIDIA RTX 6000 PRO GPU. Full implementation details are provided in the supplementary material.

\noindent\textbf{Quantitative Results.}
Table~\ref{tab:deterministic} reports deterministic ADE/FDE on ETH/UCY.
Our deterministic variant achieves the best average performance, reducing the
average FDE from $0.99$ to $0.79$ compared with the strongest prior result.
The gains are especially clear on HOTEL, UNIV, ZARA1, and ZARA2, where the
model obtains the best ADE/FDE. These scenes contain structured pedestrian
flows, group motion, and local avoidance patterns, where direction-aware
kinematic interactions provide useful cues for selecting relevant neighbors.
On ETH, where interactions are often sparser and long-range social effects are
less consistently expressed, our model ranks second in FDE and third in ADE,
while achieving the best overall average ADE/FDE across the five scenes.

\begin{table}[!h]
\centering
\caption{\textbf{Deterministic prediction results} (ADE/FDE) on ETH/UCY. Best in \textbf{bold}; second best \underline{underlined}.}
\label{tab:deterministic}
\setlength{\tabcolsep}{2pt}
\scriptsize
\begin{tabular}{lcccccc}
\toprule
Method & ETH & HOTEL & UNIV & ZARA1 & ZARA2 & Avg. \\
\midrule
Social-LSTM \cite{alahi2016social} & 1.09/2.35 & 0.79/1.76 & 0.67/1.40 & 0.47/1.00 & 0.56/1.17 & 0.72/1.54 \\
Social-GAN \cite{gupta2018social} & 1.13/2.21 & 1.01/2.18 & 0.60/1.28 & 0.42/0.91 & 0.52/1.11 & 0.74/1.54 \\
Trajectron++~\cite{salzmann2020trajectron++} & 1.02/2.00 & 0.33/0.62 & 0.53/1.19 & 0.44/0.99 & 0.32/0.73 & 0.53/1.11 \\
RSBG~\cite{sun2020recursive} & \textbf{0.80}/\textbf{1.53} & 0.33/0.64 & 0.59/1.25 & 0.40/\underline{0.86} & \underline{0.30}/\underline{0.65} & \underline{0.48}/\underline{0.99} \\
TransF~\cite{giuliari2021transformer} & 1.03/2.10 & 0.36/0.71 & 0.53/1.32 & 0.44/1.00 & 0.34/0.76 & 0.54/1.17 \\
MemoNet~\cite{xu2022remember} & 1.00/2.08 & 0.35/0.67 & 0.55/1.19 & 0.46/1.01 & 0.37/0.82 & 0.55/1.15 \\
EQMotion~\cite{xu2023eqmotion} & \underline{0.96}/1.92 & \underline{0.30}/\underline{0.58} & \underline{0.50}/\underline{1.10} & \underline{0.39}/\underline{0.86} & \underline{0.30}/0.68 & 0.49/1.03 \\
\textbf{Ours} & 0.97/\underline{1.54} & \textbf{0.19}/\textbf{0.30} & \textbf{0.43}/\textbf{0.87} & \textbf{0.30}/\textbf{0.56} & \textbf{0.29}/\textbf{0.56} & \textbf{0.45}/\textbf{0.79} \\

\bottomrule
\end{tabular}
\end{table}

\noindent\textbf{ETH/UCY Dataset.} In Table~\ref{tab:prob_all} stochastic minADE/minFDE on ETH/UCY is reported. Among explicit graph-based methods, our probabilistic
variant achieves the best result on every scene and improves the average error
from $0.25/0.44$ to $0.21/0.34$. Compared with methods beyond explicit graph
modeling, the SocialCircle variants achieve the lowest average error at
$0.17/0.27$. Our model remains competitive, tying the best minADE on HOTEL,
ZARA1, and ZARA2, while retaining a substantially lighter interaction model.
The strongest gains again appear in scenes with meaningful agent-agent motion
dependencies, supporting the idea that sparse kinematic message passing can
capture interaction modes without requiring dense pairwise reasoning.

\begin{table}[!ht]
\centering
\caption{\textbf{Probabilistic prediction results (minADE/minFDE)} on ETH/UCY.}
\label{tab:prob_all}
\scriptsize
\setlength{\tabcolsep}{2pt}

\begin{tabular}{@{}lcccccc@{}}
\toprule
Method & ETH & HOTEL & UNIV & ZARA1 & ZARA2 & Avg \\
\midrule

\multicolumn{7}{l}{\textit{GNN-based methods}} \\

Vanilla-GCN 
    & 0.74/1.21 & 0.30/0.52 & 0.55/0.96 & 0.53/0.94 & 0.40/0.71 & 0.50/0.87 \\

Social-STGCNN~\cite{mohamed2020social} 
    & 0.64/1.11 & 0.49/0.85 & 0.44/0.79 & 0.34/0.53 & 0.30/0.48 & 0.44/0.75 \\

Trajectron++~\cite{salzmann2020trajectron++} 
    & 0.59/1.14 & 0.35/0.66 & 0.54/1.13 & 0.43/0.83 & 0.43/0.85 & 0.47/0.92 \\

Social-BiGAT~\cite{kosaraju2019social} 
    & 0.81/1.52 & 0.72/1.61 & 0.60/1.26 & 0.34/0.69 & 0.42/0.84 & 0.58/1.18 \\

STGAT~\cite{huang2019stgat} 
    & 1.01/1.75 & 0.43/0.80 & 0.44/0.91 & 0.26/0.45 & 0.26/0.57 & 0.48/0.90 \\

SGCN~\cite{shi2021sgcn}
    & 0.63/1.03 & 0.32/0.55 & 0.37/0.70 & 0.29/0.53 & 0.25/0.45 & 0.37/0.65 \\

GroupNet~\cite{xu2022groupnet} 
    & \underline{0.46}/\underline{0.73} & \underline{0.15}/\underline{0.25} & \underline{0.26}/\underline{0.49} & \underline{0.21}/\underline{0.39} & \underline{0.17}/\underline{0.33} & \underline{0.25}/\underline{0.44} \\    

\textbf{Ours} 
    & \textbf{0.44}/\textbf{0.60} & \textbf{0.11}/\textbf{0.15} &
      \textbf{0.22}/\textbf{0.41} & \textbf{0.17}/\textbf{0.30} &
      \textbf{0.13}/\textbf{0.25} & \textbf{0.21/0.34} \\

\midrule
\multicolumn{7}{l}{\textit{Beyond explicit-graph methods}} \\

Social-GAN~\cite{gupta2018social} 
    & 0.87/1.62 & 0.67/1.37 & 0.76/1.52 & 0.35/0.68 & 0.42/0.84 & 0.61/1.21 \\

AgentFormer~\cite{yuan2021agentformer} 
    & 0.45/0.75 & 0.14/0.22 & 0.25/0.45 &
      \underline{0.18}/0.30 & \underline{0.14}/0.24 & 0.23/0.39 \\

TUTR~\cite{shi2023trajectory} 
    & 0.40/0.61 & \textbf{0.11}/0.18 & 0.23/0.42 &
      \underline{0.18}/0.34 & \textbf{0.13}/0.25 &
      \underline{0.21}/0.36 \\

PECNet~\cite{mangalam2020not} 
    & 0.54/0.87 & 0.18/0.24 & 0.35/0.60 &
      0.22/0.39 & 0.17/0.30 & 0.29/0.48 \\

MID~\cite{gu2022stochastic} 
    & \underline{0.39}/0.66 & 0.13/0.22 & 0.22/0.45 &
      \textbf{0.17}/0.30 & \textbf{0.13}/0.27 &
      \underline{0.21}/0.38 \\

LMTraj~\cite{bae2024can} 
    & 0.65/1.04 & 0.26/0.46 & 0.57/1.16 &
      0.51/1.01 & 0.38/0.74 & 0.48/0.88 \\

EQMotion~\cite{xu2023eqmotion} 
    & 0.40/0.61 & \underline{0.12}/0.18 & 0.23/0.43 &
      \underline{0.18}/0.32 & \textbf{0.13}/\underline{0.23} &
      \underline{0.21}/0.35 \\

LED~\cite{mao2023leapfrog} 
    & \underline{0.39}/0.58 & \textbf{0.11}/0.17 & 0.26/0.43 &
      \underline{0.18}/\textbf{0.26} &
      \textbf{0.13}/\textbf{0.22} &
      \underline{0.21}/\underline{0.33} \\

V$^2$-Net-SC~\cite{wong2024socialcircle}
    & \textbf{0.25}/\textbf{0.37} &
      \underline{0.12}/\underline{0.15} &
      \underline{0.21}/\underline{0.35} &
      \textbf{0.17}/\underline{0.29} &
      \textbf{0.13}/\textbf{0.22} &
      \textbf{0.17}/\textbf{0.27} \\

E-V$^2$-Net-SC~\cite{wong2024socialcircle}
    & \textbf{0.25}/\underline{0.38} &
      \underline{0.12}/\textbf{0.14} &
      \textbf{0.20}/\textbf{0.34} &
      \underline{0.18}/\underline{0.29} &
      \textbf{0.13}/\textbf{0.22} &
      \textbf{0.17}/\textbf{0.27} \\
      
\textbf{Ours} 
    & 0.44/0.60 &
      \textbf{0.11}/\underline{0.15} &
      0.22/0.41 &
      \textbf{0.17}/0.30 &
      \textbf{0.13}/0.25 &
      \underline{0.21}/0.34 \\
\bottomrule
\end{tabular}
\end{table}

\noindent\textbf{SDD Dataset.} Table~\ref{tab:sdd_minade} reports stochastic prediction results on the
Stanford Drone Dataset (SDD). SDD contains more crowded and visually diverse
top-view scenes than ETH/UCY, with frequent crossings, group movement, and
heterogeneous motion patterns. Our model achieves the best minADE, while the
SocialCircle variants obtain lower minFDE and overall average error. SDD places
greater emphasis on interaction modeling because the model must distinguish
dynamically relevant neighbors rather than simply aggregate nearby agents. The
improvement in minADE indicates that our method generates trajectory samples that
are closer to the ground-truth path over the full prediction horizon, while
SocialCircle achieves stronger endpoint accuracy.
\begin{table}[!h]
\centering
\caption{\textbf{Probabilistic prediction results (minADE/minFDE)} on SDD.}
\label{tab:sdd_minade}
\scriptsize
\setlength{\tabcolsep}{6pt}
\begin{tabular}{lccc}
\toprule
Method & minADE $\downarrow$ & minFDE $\downarrow$ & Avg. $\downarrow$ \\
\midrule
FLEAM~\cite{chen2022fully} & 0.67 & 1.28 & 0.98 \\
MemoNet~\cite{xu2022remember} & 0.45 & 0.64 & 0.55 \\
MANTRA~\cite{marchetti2020mantra} & 0.49 & 0.62 & 0.56 \\
BiTrap~\cite{yao2021bitrap} & 0.46 & 0.59 & 0.53 \\
MID~\cite{gu2022stochastic} & 0.38 & 0.73 & 0.56 \\

V$^2$-Net-SC~\cite{wong2024socialcircle}
    & 0.34 & \underline{0.53} & \underline{0.44} \\

E-V$^2$-Net-SC~\cite{wong2024socialcircle}
    & \underline{0.33} & \textbf{0.52} & \textbf{0.43} \\

\midrule
\textbf{Ours} & \textbf{0.32} & 0.62 & 0.47 \\
\bottomrule
\end{tabular}
\end{table}

\noindent\textbf{NBA Dataset.} Table~\ref{tab:nba} reports minADE/minFDE on the NBA SportVU dataset across
$1/2/3/4$\,s prediction horizons. NBA is a highly interactive setting with
fast motion behavior and coordinated group dynamics among players and the ball.
Our method is consistently among the top two methods across all horizons,
obtaining the best ADE at $2.0$\,s. The gap to the best method is small at
longer horizons, where uncertainty accumulates and coordinated basketball plays
become harder to disambiguate. In this dense $11$-agent setting, $k=10$ is
an upper bound because kinematic gating precedes top-$k$ selection; MEDIG
retains $78.5/110$ directed edges on average (density $0.714$). NBA therefore
uses a denser, dataset-specific connectivity regime, with detailed sensitivity
and transfer analyses provided in the supplementary material.

\begin{table}[h]
\centering
\caption{\textbf{Probabilistic prediction results (minADE/minFDE)} on the NBA SportVU, reported at $1/2/3/4$\,s horizons.
}
\label{tab:nba}
\scriptsize
\setlength{\tabcolsep}{5.5pt}
\begin{tabular}{lcccc}
\toprule
Method & $1.0$\,s & $2.0$\,s & $3.0$\,s & $4.0$\,s \\
\midrule
Social-GAN~\cite{gupta2018social}        & 0.41/0.62 & 0.81/1.32 & 1.19/1.94 & 1.59/2.41 \\
PECNet~\cite{mangalam2020not}               & 0.40/0.71 & 0.83/1.61 & 1.27/2.44 & 1.69/2.95 \\
NPSN~\cite{bae2022non}                   & 0.35/0.58 & 0.68/1.23 & 1.01/1.76 & 1.31/1.79 \\
MemoNet~\cite{xu2022remember}             & 0.38/0.56 & 0.71/1.14 & 1.00/1.57 & 1.25/1.47 \\
STAR~\cite{YuMa2020Spatio}                   & 0.43/0.66 & 0.75/1.24 & 1.03/1.51 & 1.13/2.01 \\
GroupNet~\cite{xu2022groupnet}           & 0.26/0.34 & 0.49/0.70 & 0.73/1.02 & 0.96/1.30 \\
MID~\cite{gu2022stochastic}                     & 0.28/0.37 & 0.51/0.72 & 0.71/0.98 & 0.96/1.27 \\
V$^2$-Net-SC~\cite{wong2024socialcircle}
    & -- & 0.67/0.92 & -- & 1.22/1.51 \\

E-V$^2$-Net-SC~\cite{wong2024socialcircle}
    & -- & 0.67/0.90 & -- & 1.18/1.46 \\
LED~\cite{mao2023leapfrog}                     & \textbf{0.18}/\textbf{0.27} & \underline{0.37}/\textbf{0.56} & \textbf{0.58}/\textbf{0.84} & \textbf{0.81}/\textbf{1.10} \\
\textbf{Ours}              & \underline{0.20}/\underline{0.31} & \textbf{0.34}/\underline{0.62} & \underline{0.59}/\underline{0.96} & \underline{0.84}/\underline{1.16} \\
\bottomrule
\end{tabular}
\end{table}


\noindent\textbf{Model Efficiency Comparison.} In Table~\ref{tab:params}, we report parameter count, computational cost, inference latency, and training time, with latency and epoch time measured on ETH/UCY. Our method is the most compact model, using 0.16M parameters, which is 2.8$\times$ fewer than TUTR, approximately 12$\times$ fewer than E-V$^2$-Net-SC, and over 60$\times$ fewer than MID and LED. With 32.9M FLOPs, it also requires approximately 12$\times$ less computation than E-V$^2$-Net-SC and 13$\times$ less than TUTR/EQMotion. Our method outperforms E-V$^2$-Net-SC across all reported efficiency metrics, reducing latency from 7.39 to 3.15\,ms and epoch time from 2.94 to 0.52\,s, while retaining the second-best latency and fastest training overall. Unlike SocialCircle's angular feature augmentation, MEDIG uses sparse kinematics-conditioned message passing with continuous angular kernels, yielding a substantially lighter and faster predictor while maintaining competitive accuracy across ETH/UCY, SDD, and NBA.

\begin{table}[h]
\centering
\caption{\textbf{Model efficiency comparison.}}
\label{tab:params}
\setlength{\tabcolsep}{5pt}
\scriptsize
\begin{tabular}{lcccc}
\toprule
Method & Params (M) & FLOPs & Latency (ms) & Epoch Time (s) \\
\midrule
AgentFormer~\cite{yuan2021agentformer} & 7.09 & 114.7G & 47.39 & 333.3 \\
EQMotion~\cite{xu2023eqmotion} & 3.03 & 419.2M & 28.6 & 31.3 \\
TUTR~\cite{shi2023trajectory} & \underline{0.44} & 424.3M & \textbf{1.66} & 3.3 \\
MID~\cite{gu2022stochastic} & 9.65 & 3762.5G & 187.0 & 210.1 \\
PECNet~\cite{mangalam2020not} & 2.10 & 621.2M & 10.76 & 3.2 \\
LED~\cite{mao2023leapfrog} & 10.87 & 97.4G & 28.98 & 29.8 \\
E-V$^2$-Net-SC~\cite{wong2024socialcircle}
    & 1.99 & \underline{393.7M} & 7.39 & \underline{2.94} \\
\textbf{Ours} & \textbf{0.16} & \textbf{32.9M} & \underline{3.15} & \textbf{0.52} \\
\bottomrule
\end{tabular}
\end{table}

\begin{figure*}
\centering
\captionsetup[subfigure]{justification=centering, font=small}

\begin{subfigure}[b]{0.23\textwidth}
    \includegraphics[width=\linewidth]{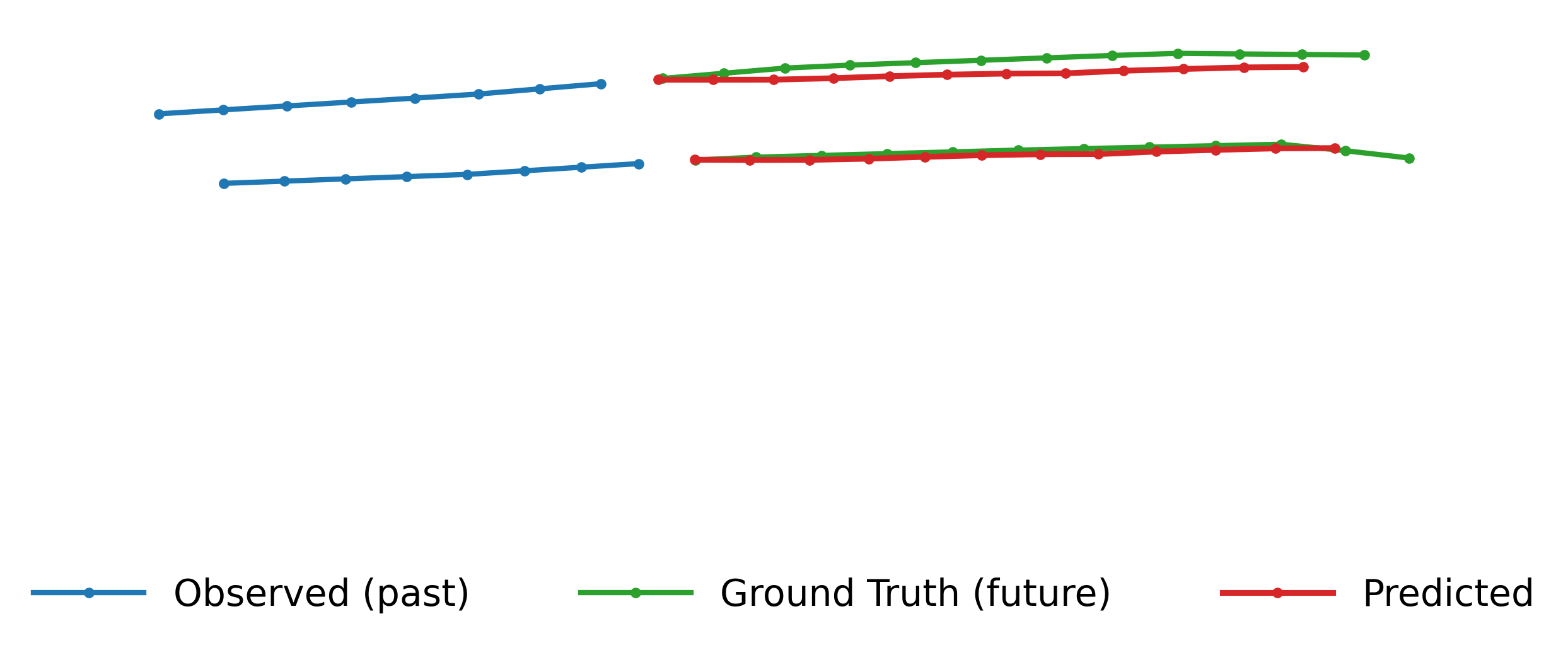}
    \caption{Following behavior}
\end{subfigure}\hfill
\begin{subfigure}[b]{0.23\textwidth}
    \includegraphics[width=\linewidth]{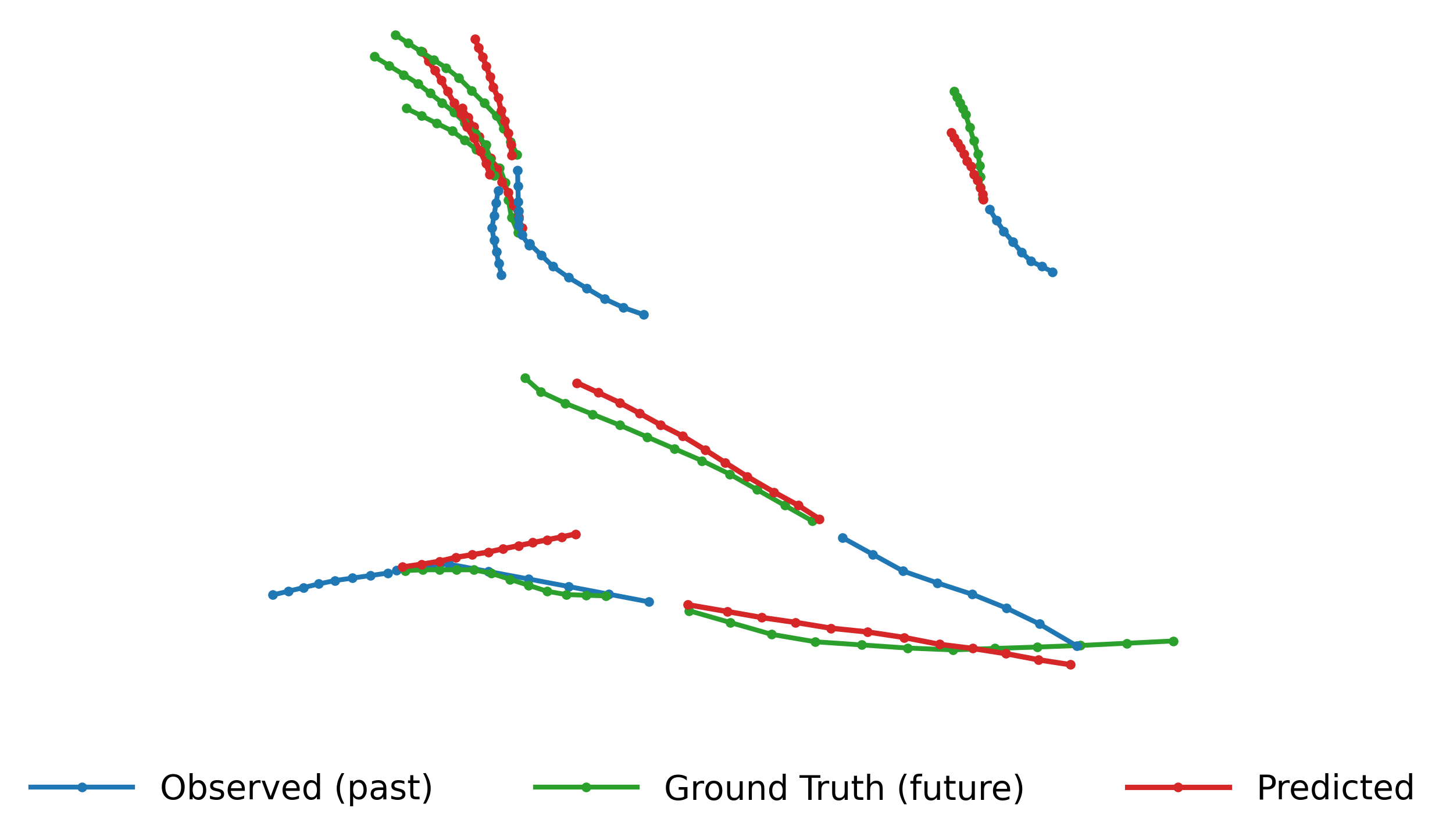}
    \caption{Avoidance maneuver}
\end{subfigure}\hfill
\begin{subfigure}[b]{0.23\textwidth}
    \includegraphics[width=\linewidth]{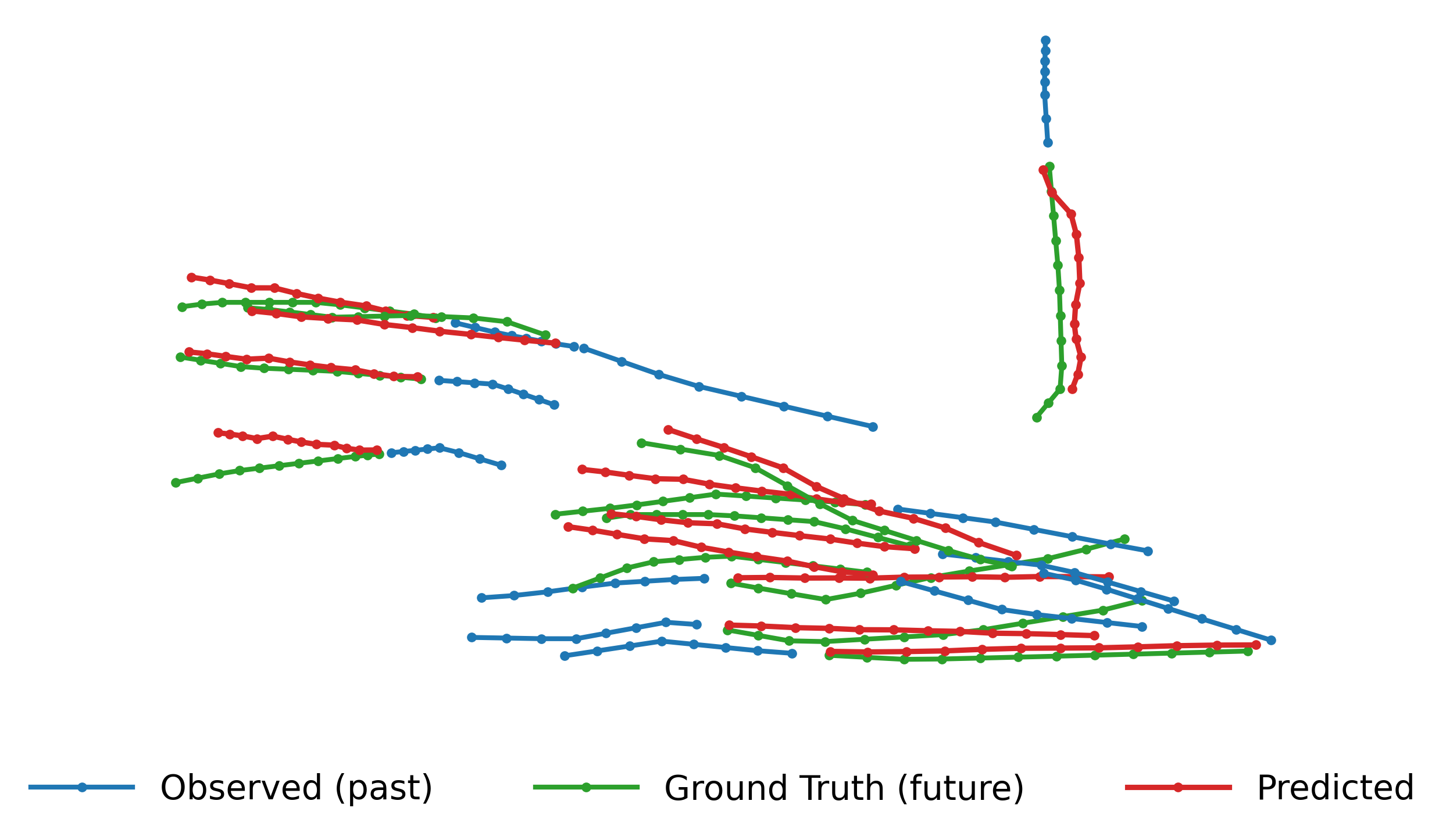}
    \caption{Group merging}
\end{subfigure}\hfill
\begin{subfigure}[b]{0.23\textwidth}
    \includegraphics[width=\linewidth]{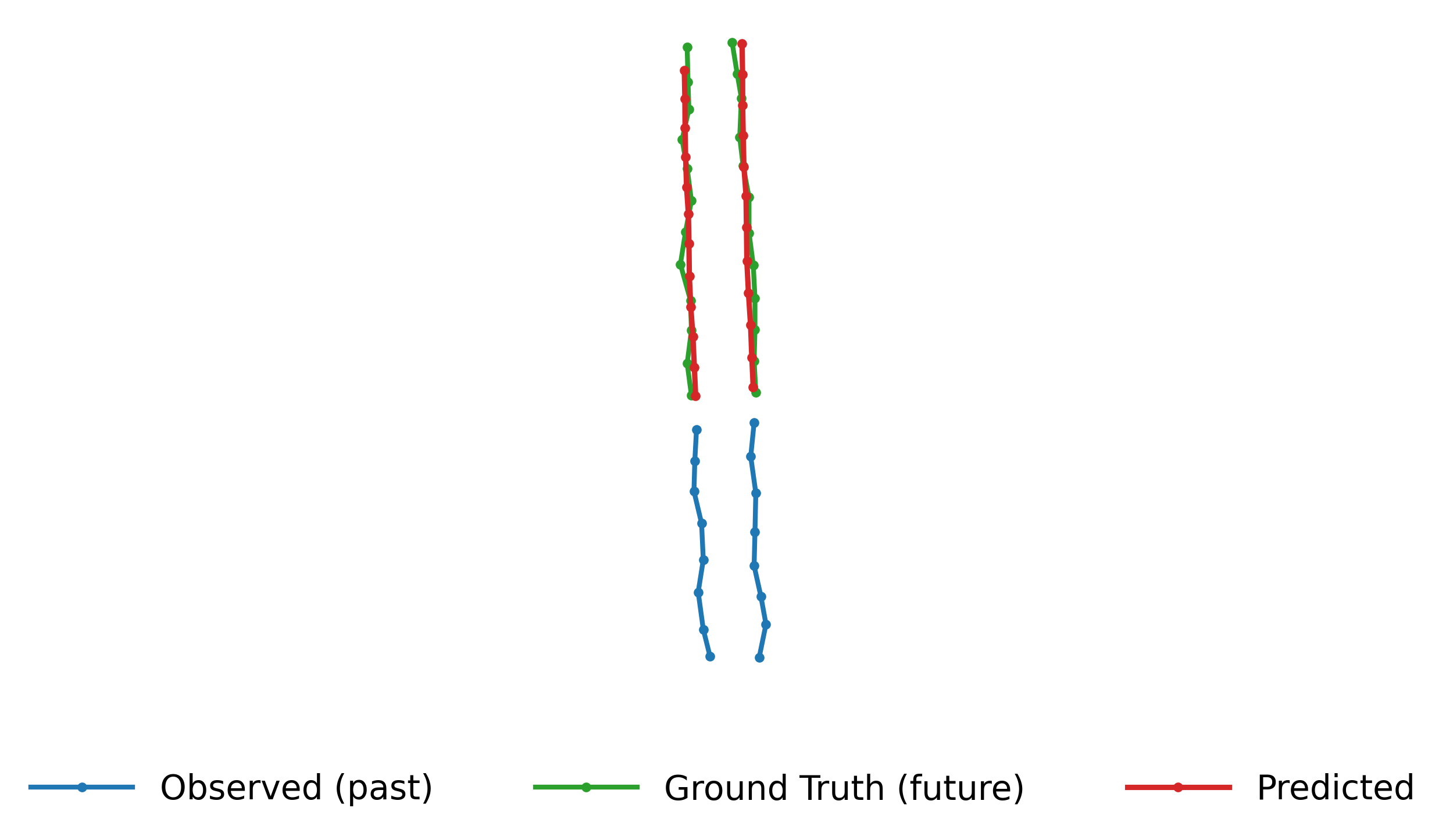}
    \caption{Smooth trajectory}
\end{subfigure}


\begin{subfigure}[b]{0.3\textwidth}
    \includegraphics[width=\linewidth]{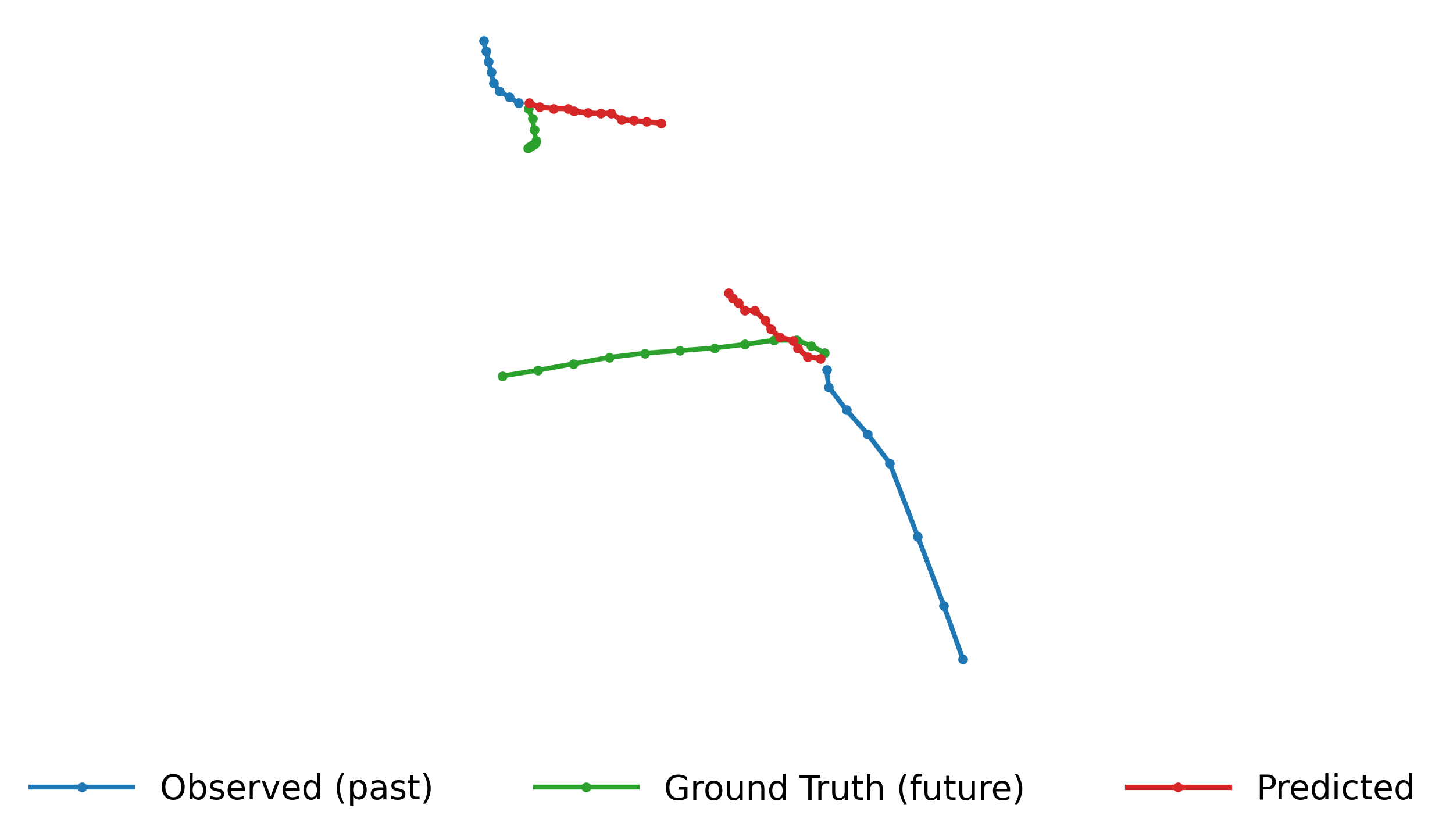}
    \caption{Abrupt direction change}
\end{subfigure}\hfill
\begin{subfigure}[b]{0.3\textwidth}
    \includegraphics[width=\linewidth]{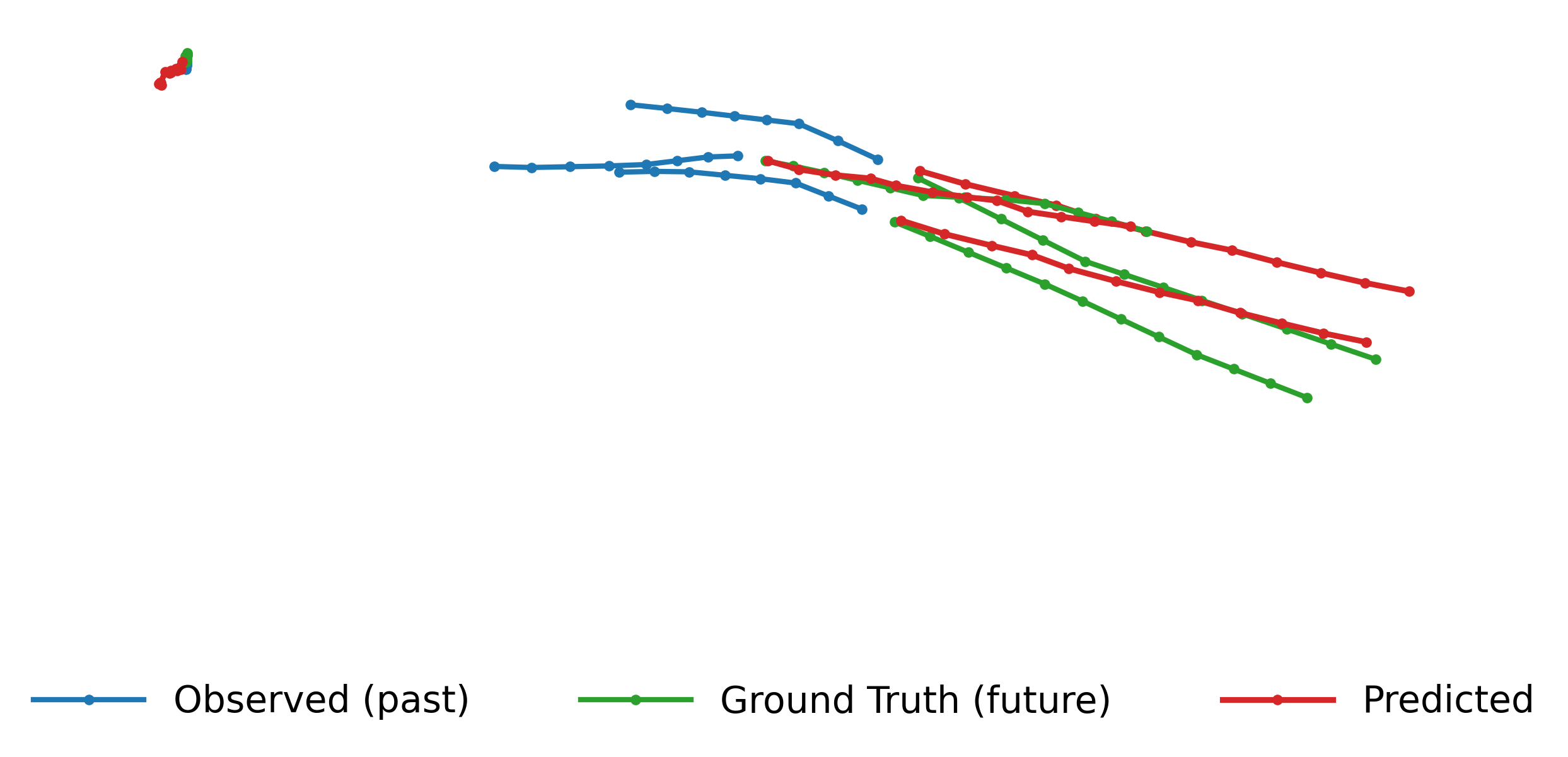}
    \caption{Late interaction response}
\end{subfigure}\hfill
\begin{subfigure}[b]{0.3\textwidth}
    \includegraphics[width=\linewidth]{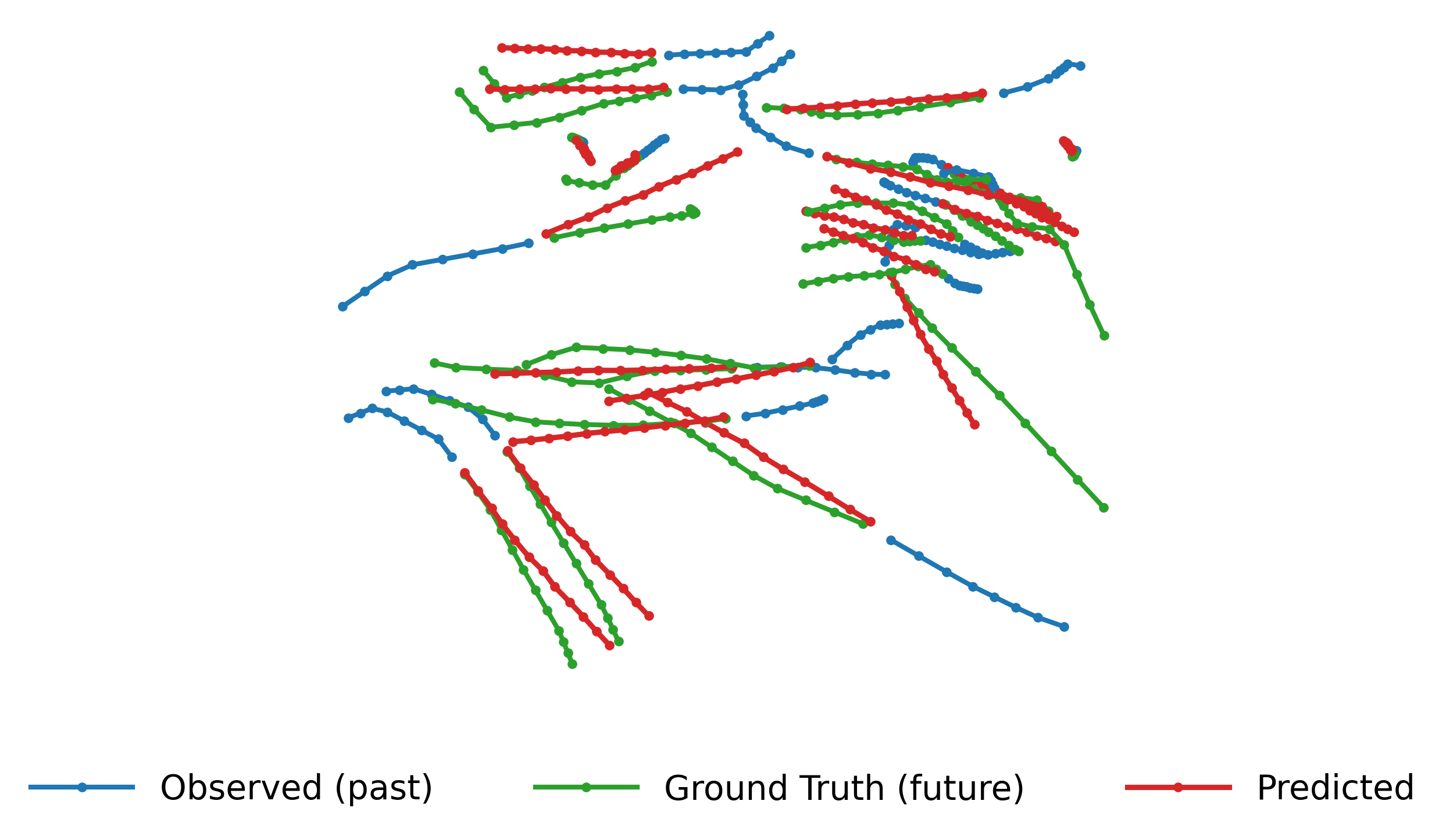}
    \caption{Occlusion-induced error}
\end{subfigure}


\caption{
Qualitative results on ETH and UCY datasets. 
\textbf{Top:} Successful predictions. Observed trajectories (\textcolor{blue}{blue}), predicted (\textcolor{green!60!black}{green}), and ground truth (\textcolor{red}{red}). 
The model accurately captures merging, following, and avoidance behaviors. 
\textbf{Bottom:} Representative failure cases in highly congested or abrupt-motion scenarios, where predictions deviate due to interaction ambiguity or occlusion.
}
\label{fig:qualitative_predictions}
\end{figure*}

\begin{figure}[!h]
\centering
\captionsetup[subfigure]{justification=centering}

\begin{subfigure}[t]{0.45\linewidth}
    \includegraphics[width=\linewidth]{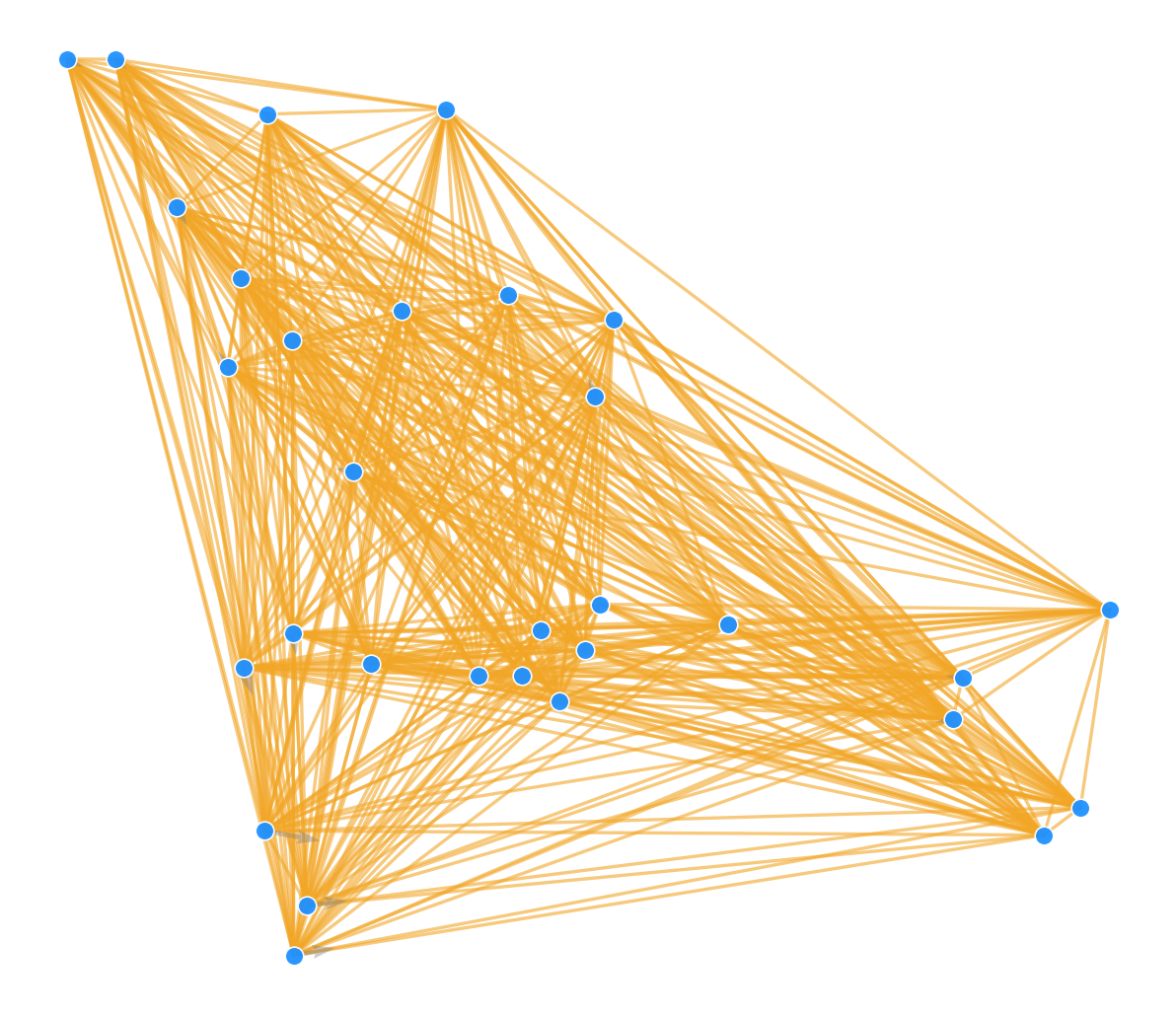}
    \caption{Fully connected (Crowded Scene)}
\end{subfigure}
\hfill
\begin{subfigure}[t]{0.45\linewidth}
    \includegraphics[width=\linewidth]{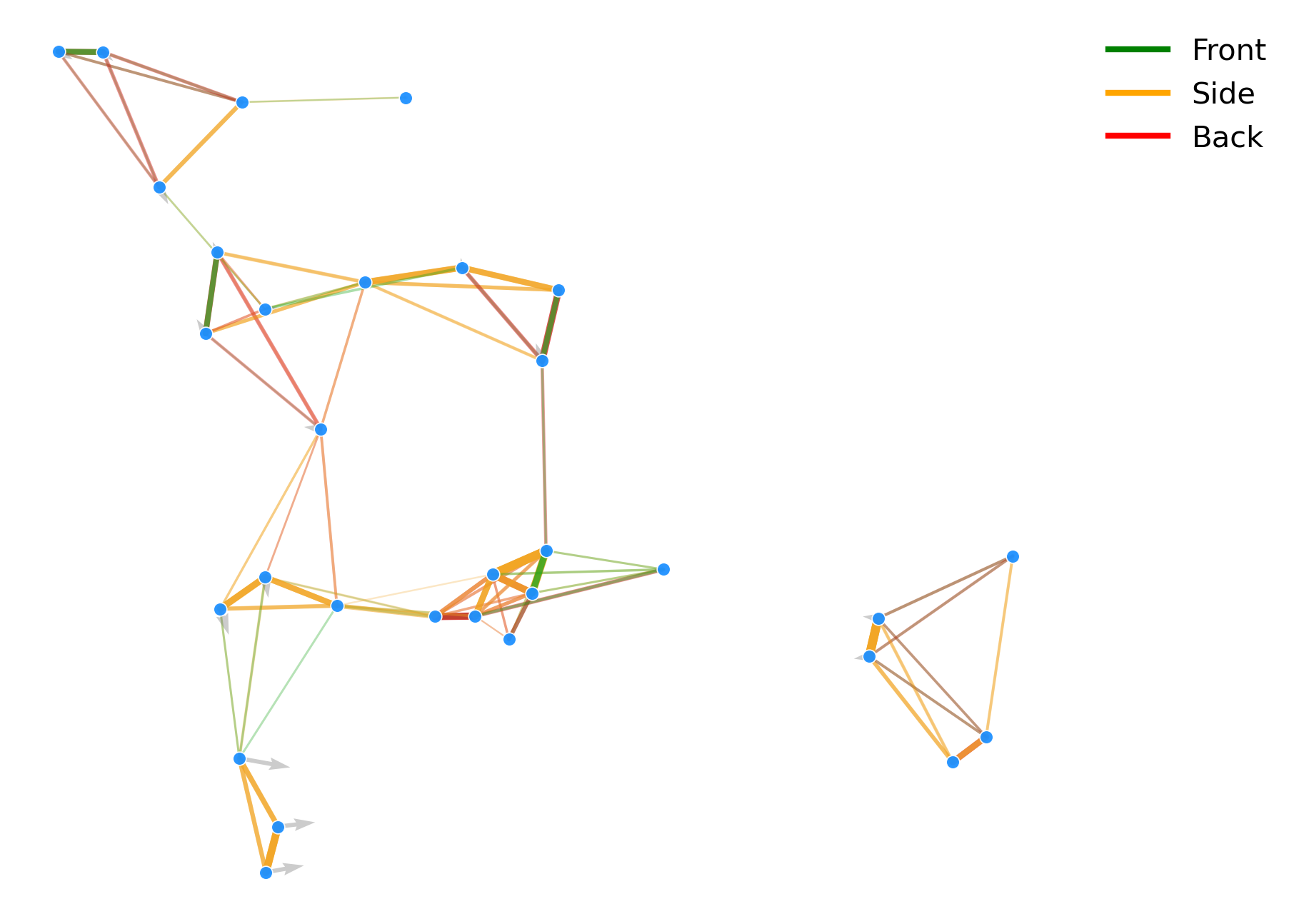}
    \caption{MEDIG (Crowded Scene)}
\end{subfigure}


\begin{subfigure}[t]{0.45\linewidth}
    \includegraphics[width=\linewidth]{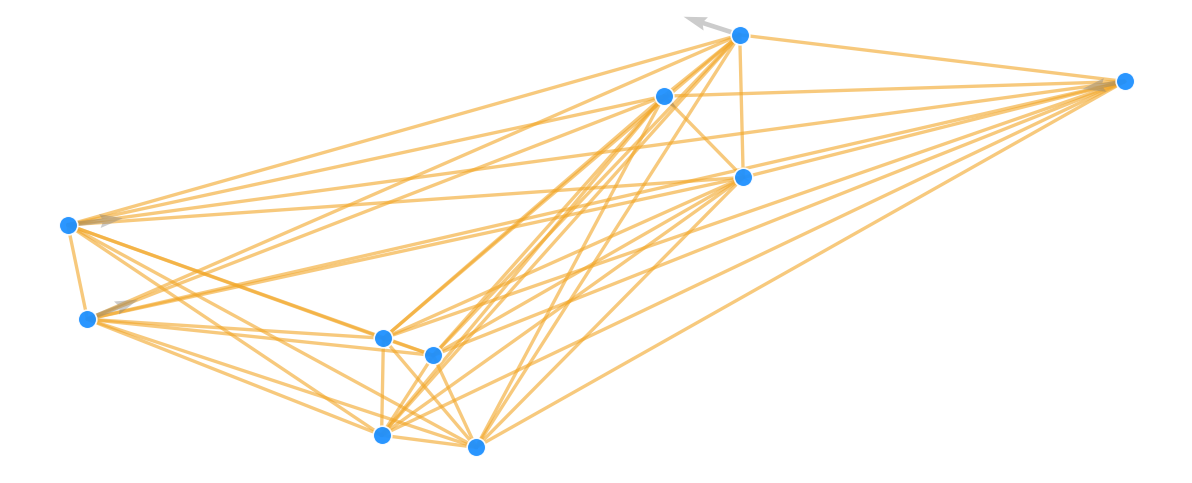}
    \caption{Fully connected (Sparse Scene)}
\end{subfigure}
\hfill
\begin{subfigure}[t]{0.45\linewidth}
    \includegraphics[width=\linewidth]{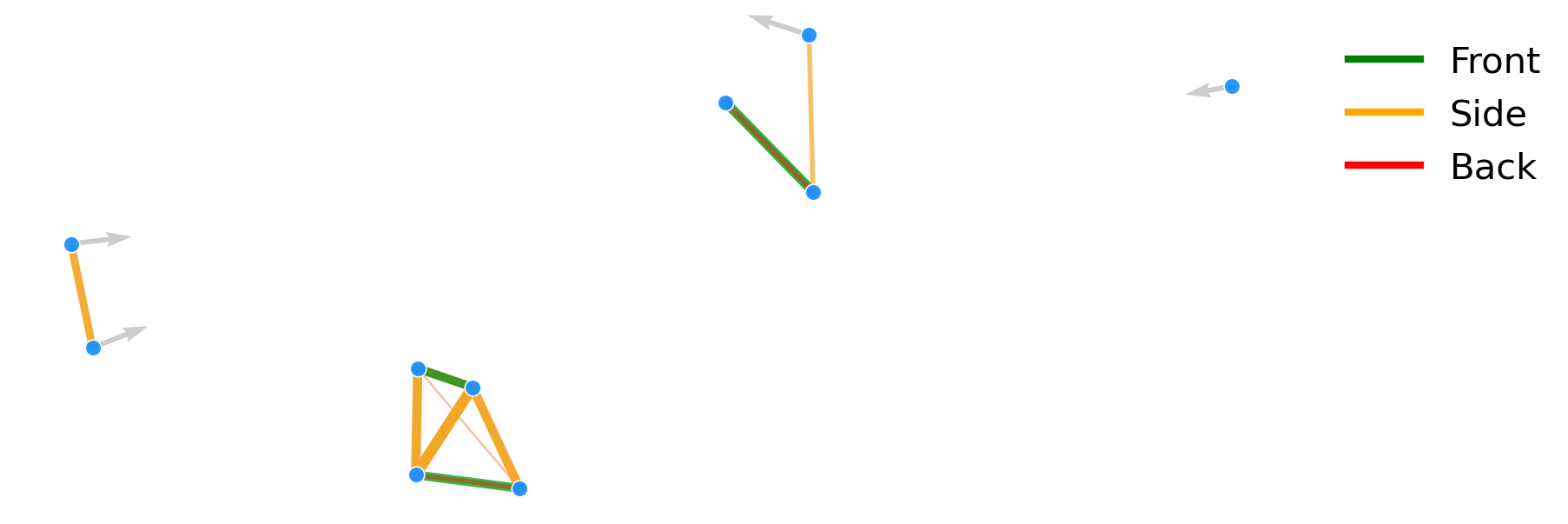}
    \caption{MEDIG (Sparse Scene)}
\end{subfigure}


\caption{
Comparison of graph connectivity patterns for two different scenes.
Fully connected graphs densely link all agents, whereas MEDIG produces sparse,
direction-aware edges that align with motion dynamics and scene structure.
}
\label{fig:qualitative_graphs}
\end{figure}

\noindent\textbf{Qualitative Results.} 
Figure~\ref{fig:qualitative_predictions} shows qualitative prediction results on ETH and UCY scenes.
Observed trajectories are shown in blue, predicted trajectories in green, and ground truth in red. 
Our model successfully anticipates merging, following, and avoidance behaviors, while most failure cases occur in highly congested or abrupt-motion scenarios.
Figure~\ref{fig:qualitative_graphs} compares graph topologies produced by MEDIG and a fully connected graph, illustrating the emergence of sparse and directional connectivity.

\noindent\textbf{Ablation Studies.} Table~\ref{tab:ablation_model} analyzes each architectural component. Without FiLM modulation, the model reduces to uniform message passing, losing edge-specific adaptation and becoming more sensitive to irrelevant neighbors. Removing directional kernels makes interactions isotropic, weakening distinctions among front-facing, lateral, and trailing agents. The largest degradation occurs without kinematic conditioning, as the model loses cues such as approach rate and TCA and reverts toward correlation-driven aggregation. Removing the global scene-conditioning gate reduces robustness in dense or heterogeneous scenes, where global flow and crowd layout shape interactions.

Table~\ref{tab:ablation_graph} compares graph construction strategies. Fully connected graphs introduce irrelevant neighbors and high message-passing cost, while proximity and attention-based graphs either rely only on distance or require dense pairwise scoring. MEDIG reduces the average number of edges from $264.0$ to $75.7$ while achieving the best predictive accuracy. Runtime includes graph construction (batch size $1$; Table~\ref{tab:params}
uses batch size $8$); for these lightweight variants, this overhead offsets much
of the computational benefit of sparser message passing.

\begin{table}[!h]
\centering
\caption{\textbf{Component ablation study} averaged over ETH/UCY.}
\label{tab:ablation_model}
\setlength{\tabcolsep}{4pt}
\scriptsize
\begin{tabular}{c c c c | r r}
\toprule
\textbf{Kinematics} & \textbf{Anisotropy} & \textbf{Gate} & \textbf{FiLM-modulation} & \textbf{minADE} & \textbf{minFDE} \\
\midrule
$\times$ & $\times$ & $\times$ & $\times$ & 0.74 & 1.21 \\ 
$\checkmark$ & $\times$ & $\times$ & $\times$ & 0.50 & 0.87 \\
$\times$ & $\checkmark$ & $\times$ & $\times$ & 0.55 & 0.80 \\
$\checkmark$ & $\times$ & $\checkmark$ & $\times$ & 0.46 & 0.75 \\
$\checkmark$ & $\checkmark$ & $\times$ & $\times$ & 0.34 & 0.50 \\ 
$\checkmark$ & $\times$ & $\times$ & $\checkmark$ & 0.34 & 0.52 \\

$\checkmark$ & $\checkmark$ & $\checkmark$ & $\times$ & 0.32 & 0.49 \\
$\checkmark$ & $\times$ & $\checkmark$ & $\checkmark$ & 0.27 & 0.44 \\ 
$\checkmark$ & $\checkmark$ & $\times$ & $\checkmark$ & \underline{0.26} & \underline{0.40} \\
$\times$ & $\checkmark$ & $\checkmark$ & $\checkmark$ & 0.33 & 0.51 \\
$\checkmark$ & $\checkmark$ & $\checkmark$ & $\checkmark$ & \textbf{0.21} & \textbf{0.34} \\ 
\bottomrule
\end{tabular}
\end{table}

\begin{table}[!h]
\centering
\caption{\textbf{Graph construction ablation.} Trade-off between accuracy, sparsity, and efficiency across ETH/UCY.}
\label{tab:ablation_graph}
\setlength{\tabcolsep}{4pt}
\scriptsize
\begin{tabular}{lcccc}
\toprule
Graph type & ADE/FDE & AvgEdges & EdgeDensity & Runtime (s) \\
\midrule
Full Graph & 0.32/0.46 & 264.0 & 1.00 & \underline{0.026} \\
Proximity Graph & 0.37/0.58 & \underline{84.9} & \underline{0.32} & \textbf{0.025} \\
Attention Graph & \underline{0.24}/\underline{0.38} & 155.0 & 0.59 & 0.031 \\
\textbf{MEDIG (ours)} & \textbf{0.21}/\textbf{0.34} & \textbf{75.7} & \textbf{0.29} &0.027 \\
\bottomrule
\end{tabular}
\end{table}

%% file: sec/5_conclusion.tex
\section{Conclusion}
We introduced a kinematics-conditioned graph hypernetwork that models human interactions through direction-aware message passing. Unlike fully connected graphs or attention-driven predictors, we construct sparse interaction graphs directly from measurable motion cues like relative velocity, alignment, distance attenuation, and TCA, yielding a connectivity pattern that aligns with how real pedestrians influence one another. A FiLM-modulated hypernetwork further adapts message flow using both local relations and global crowd dynamics, enabling continuous anisotropy without handcrafted partitioning.

Across deterministic and probabilistic settings, our approach outperforms graph-based baselines. We deliver performance competitive with heavier architectures while maintaining significantly lower computational complexity. Ablations confirm that each component contributes meaningfully to overall accuracy.

Our results demonstrate that embedding physically meaningful structure into graph neural predictors is an effective and lightweight alternative to dense attention or fully learned graph formation. Future work will extend beyond pairwise relations to capture higher-order and group-level interactions in dense multi-agent scenes. We plan to incorporate semantic scene context, such as maps and static obstacles, into the kinematic descriptors for structured environments including autonomous driving. Finally, we aim to improve long-horizon prediction while preserving the model's computational efficiency.